\documentclass{article}

\usepackage[english]{babel}

\usepackage[letterpaper,top=2cm,bottom=2cm,left=3cm,right=3cm,marginparwidth=1.75cm]{geometry}

\usepackage{amsmath,amssymb}
\usepackage{graphicx}
\usepackage{booktabs}
\usepackage{comment}
\usepackage{tikz}
\usetikzlibrary{arrows.meta,positioning,fit}
\usepackage{algorithm}
\usepackage{algpseudocode}
\usepackage[colorlinks=true, allcolors=blue]{hyperref}
\usepackage{xurl}
\usepackage{array}

\title{AlphaDefect: Expert-Level AI for Slay the Spire 2}
\author{
  Qiyue Chen\thanks{Corresponding author. Author contributions are given at the end of the main text, before the references.}\\{\small AgroParisTech, Université Paris-Saclay}\\{\small\texttt{qiyue.chen@agroparistech.fr}}
  \and
  Taian Yu\\{\small The University of New South Wales}\\{\small\texttt{iryougi@iryougi.com}}
  \and
  Junpei Komiyama\\{\small Mohamed bin Zayed University of Artificial Intelligence}\\{\small\texttt{junpei@komiyama.info}}
} 

\begin{document}
\maketitle

\begin{abstract}
Slay the Spire 2 (STS2) is a single-player deck-building game that requires long-term planning under heavy randomness.
We present AlphaDefect, an expert-level AI system for STS2: for each character, one network trained with proximal policy optimization (PPO) makes every decision of a run, in combat and outside it.
At the highest difficulty, Ascension~10, on two versions of the game current in November 2026 and on test seeds not played in training, AlphaDefect wins 62--77\% of runs across six characters with greedy play and 68--80\% with test-time search, after 1.3 to 6.7 billion decisions of training per character; runs uploaded by human players of all levels are won 11--20\% of the time at the same difficulty.
The networks read the game as entity records that identify each card, relic and enemy through a learned embedding and attributes from the game data, together with consequences of public information computed by the engine.
Ablations show how much the agent depends on this encoding: without these identities, the network did not win a single run, and without the computed consequences it won 8.3 points fewer.
To our knowledge, AlphaDefect is the strongest publicly reported STS2 agent as of November 2026.
\end{abstract}

\section{Introduction}

\begin{figure}[t]
\centering
\includegraphics[width=\linewidth]{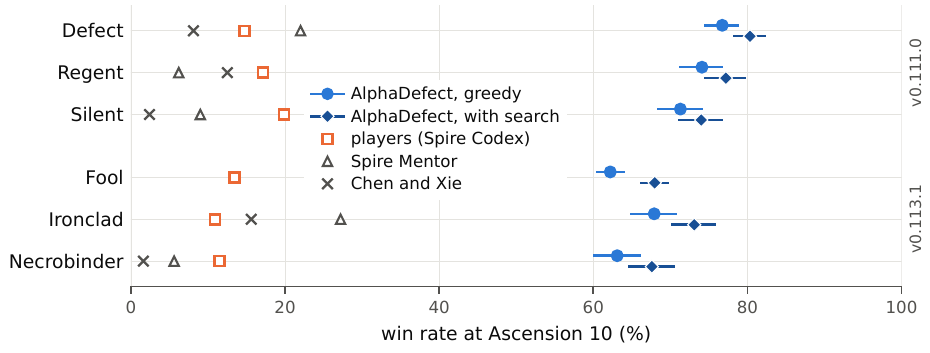}
\caption{Win rates at Ascension~10 by character.
AlphaDefect: greedy play and play with test-time search on test seeds not played in training (Table~\ref{tab:main-results}), with 95\% Wilson intervals.
Players: single-player runs uploaded to Spire Codex by players of all levels of skill, on the same game version as the AlphaDefect result~\cite{spirecodex}.
Other agents: the search-free policies of Chen and Xie on v0.107.1~\cite{chen2026zerosearch} and the October~2026 tree-search measurements reported by Spire Mentor's author~\cite{spirementor}, on an engine version that, according to Chen and Xie, still granted rewards after the first Act-3 boss.
The labels on the right give the game versions of the AlphaDefect results; game versions, seeds and protocols differ between the sources (Section~\ref{sec:comparison}).}
\label{fig:headline}
\end{figure}

Games have long served as milestones for artificial intelligence.
In chess, IBM's Deep Blue defeated the world champion Garry Kasparov in a six-game match in 1997, combining massively parallel search with hand-crafted evaluation~\cite{campbell2002deepblue}.
Go was regarded as the most challenging classic board game because of its enormous search space and the difficulty of evaluating positions, and AlphaGo defeated a professional player, the European champion, a feat previously thought to be at least a decade away~\cite{silver2016alphago}.
Its successors learned without human data~\cite{silver2017alphagozero} and mastered chess, shogi, and Go with a single algorithm~\cite{silver2018alphazero}.
Subsequent systems reached expert level in games with imperfect information, such as poker~\cite{brown2018libratus,brown2019pluribus}, and in video games with long horizons and large state and action spaces, such as Dota~2~\cite{berner2019dota} and StarCraft~II~\cite{vinyals2019alphastar}.
AlphaZero, OpenAI Five, and AlphaStar learn through large-scale, distributed self-play~\cite{silver2018alphazero,berner2019dota,vinyals2019alphastar}.

Slay the Spire 2 (STS2) is a deck-building roguelike, a genre in which these difficulties appear in a different combination.
The game has attracted wide attention: its developer reported 3 million copies sold in the first week after its early-access release in March 2026~\cite{gamesradar2026sts2sales}, and a market research firm estimated Steam sales at about 7.9 million copies by November 2026~\cite{techpowerup2026sts2sales}.
A run consists of several hundred decisions: the player fights turn-based battles, adds cards, relics, and potions to the deck and inventory, and chooses a route through procedurally generated maps.
Three properties distinguish STS2 from the games above.
First, it is a single-player game.
The opponent is not another player but the game itself, so the agent cannot rely on the self-play that drove the successes above, and its strength is measured by its win rate rather than by its results against other players.
Second, the game is highly random.
The cards offered, the cards drawn, the enemies, and the events all depend on the seed, so a run can be won after poor decisions and lost after good ones.
Even a single fight with a fixed deck is largely decided by the order in which cards are drawn.
Much of this randomness is also hidden: the player cannot see the order of the draw pile, the upcoming encounters, the random branches of enemy behavior or the state of the random number generators, so unlike in Go or chess the future cannot be explored by searching a known game tree.
The outcome of a single run is therefore weak evidence about the quality of the decisions in it, both for a learning agent and for evaluating it.
Third, the reward is extremely sparse. The strategy that leads to a higher short-term survival rate (i.e., picks more cards) often negatively affects the long-term success of the run.
For example, the final bosses have large HP and require a deck whose damage keeps growing over a fight, which is often very different from the deck that does well early.
Ultimately, the only feedback is whether the whole run was won or lost, and the effect of an early decision, such as adding a card on the second floor, is spread over every later fight.
The decisions that matter most are also rare: for an Ironclad policy at Ascension~0, the Act-3 boss accounted for 7\% of its decisions but 59\% of its losses~\cite{chen2026zerosearch}.
Moreover, the agent shapes its own action space, since the cards it adds to its deck determine which actions are available in later fights~\cite{lowe2026spirebench}.
Finally, a run interleaves two kinds of decisions: turn-level combat, a combinatorial planning problem, and run-level choices such as card picks, shops, routes and events, each with its own action space and timescale (Section~\ref{sec:learning-problem}).

This combination is also common outside games.
Many high-value applications of AI, such as logistics, portfolio construction, clinical treatment planning, and autonomous agents in general, require a sequence of hundreds of interdependent decisions under uncertainty.
In such problems a single decision maker acts against an uncertain environment rather than an adversary, a single mistake can ruin the outcome, outcomes are noisy even under good decisions, and feedback often arrives only at the very end.
STS2 offers a controlled, reproducible, and inexpensive setting in which to study learning under these conditions, with a large population of human players as a reference.
Methods that succeed here should also carry over to other stochastic single-player games with long horizons and heterogeneous decisions, such as Minecraft~\cite{wang2023voyager} and Civilization~VI~\cite{andrews2026civbench}.

These properties make STS2 hard for current AI systems.
Large language models match the average human player in the original game but remain far behind experts~\cite{lowe2026spirebench}.
On STS2, the best frontier models win up to 35\% of runs at the highest difficulty of v0.107.1 through a computer-use harness~\cite{ou2026epochsts}, and an agent that supplies a language model with extensive game knowledge wins 6 of 10 runs at the lowest difficulty~\cite{cheng2026agenticsts}.
Agents trained specifically for STS2 with reinforcement learning, with or without tree search, win only a small fraction of runs at the highest difficulty, Ascension~10 (A10).
On an earlier version of the game, the best published search-free policies win 8\% of runs averaged over five characters~\cite{chen2026zerosearch}; Spire Mentor's October~2026 measurements report a five-character mean of 14\% with tree search~\cite{spirementor}, or 8.8\% when run by Chen and Xie in their own host~\cite{chen2026zerosearch}.
Beyond being a benchmark, strong agents for such games are useful in practice, for example to measure difficulty during game development~\cite{xiao2024testers}, and a policy that needs no search at decision time can run inside the game on a CPU~\cite{chen2026zerosearch}.

We present AlphaDefect, a reinforcement learning system that plays STS2 at Ascension~10.
For each character, one network makes every decision of a run, in combat and outside it, by scoring each legal action, or candidate, that the game offers.
The network reads the game as a set of entity records (the player, enemies, cards, relics and so on), each with a learned embedding of its ID, a row of attributes from the game data and numerical features, among them consequences of public information that the engine computes, such as the damage of each card after all modifiers.
In ablations on the Ironclad, the network did not win a single run without the identities of cards, relics and enemies, and won 8.3 points fewer without the engine-computed consequences (Section~\ref{sec:ablation}).
It is trained with proximal policy optimization (PPO)~\cite{schulman2017ppo} against the game's own rules, which we run headlessly, and can learn without any human games or demonstrations: none of the evaluated networks was trained on human games\footnote{The only human data that entered their training was an early reward term in the Defect's training, which compared the size of the deck with that of winning human decks; it was removed during that training. For more explicit use of human data, see Appendix~\ref{sec:human-run-data}.}, and Appendix~\ref{sec:human-run-data} reports a separate variant that starts combat practice from states of human runs.
At decision time the network can play alone, choosing the most probable candidate at every decision (greedy play).
In the hardest fights, it can also guide a search that is designed to use only information available to the player.

With greedy play on v0.111.0, the version current in early November 2026, the networks win 76.7\%, 74.1\% and 71.3\% of runs with the Defect, the Regent and the Silent, and 80.3\%, 77.2\% and 74.0\% with search in the two boss fights that end the run (the Act-3 bosses, Section~\ref{sec:background}).
Runs uploaded by human players of all levels with these three characters on v0.111.0 at the same difficulty are won 15--20\% of the time~\cite{spirecodex}.
Figure~\ref{fig:headline} sets these results beside community statistics and other agents.
On v0.113.1, the Fool wins 62.2\% greedily and 68.0\% with search in all boss and elite fights, the Ironclad 67.9\% and 73.1\%, and the Necrobinder 63.1\% and 67.6\%.
These win rates, measured on test seeds never played in training (Section~\ref{sec:eval-protocol}), carry over to the released game: in 866 runs played on a public livestream\footnote{\url{https://live.bilibili.com/26239602} (Bilibili).}, the networks won 645, against 642.5 expected.
To our knowledge, these are the highest publicly reported win rates of any AI agent on STS2 at Ascension~10 as of November 2026.

\subsection{Background}

\label{sec:background}

\begin{figure}[t]
  \centering
  \includegraphics[width=0.49\linewidth]{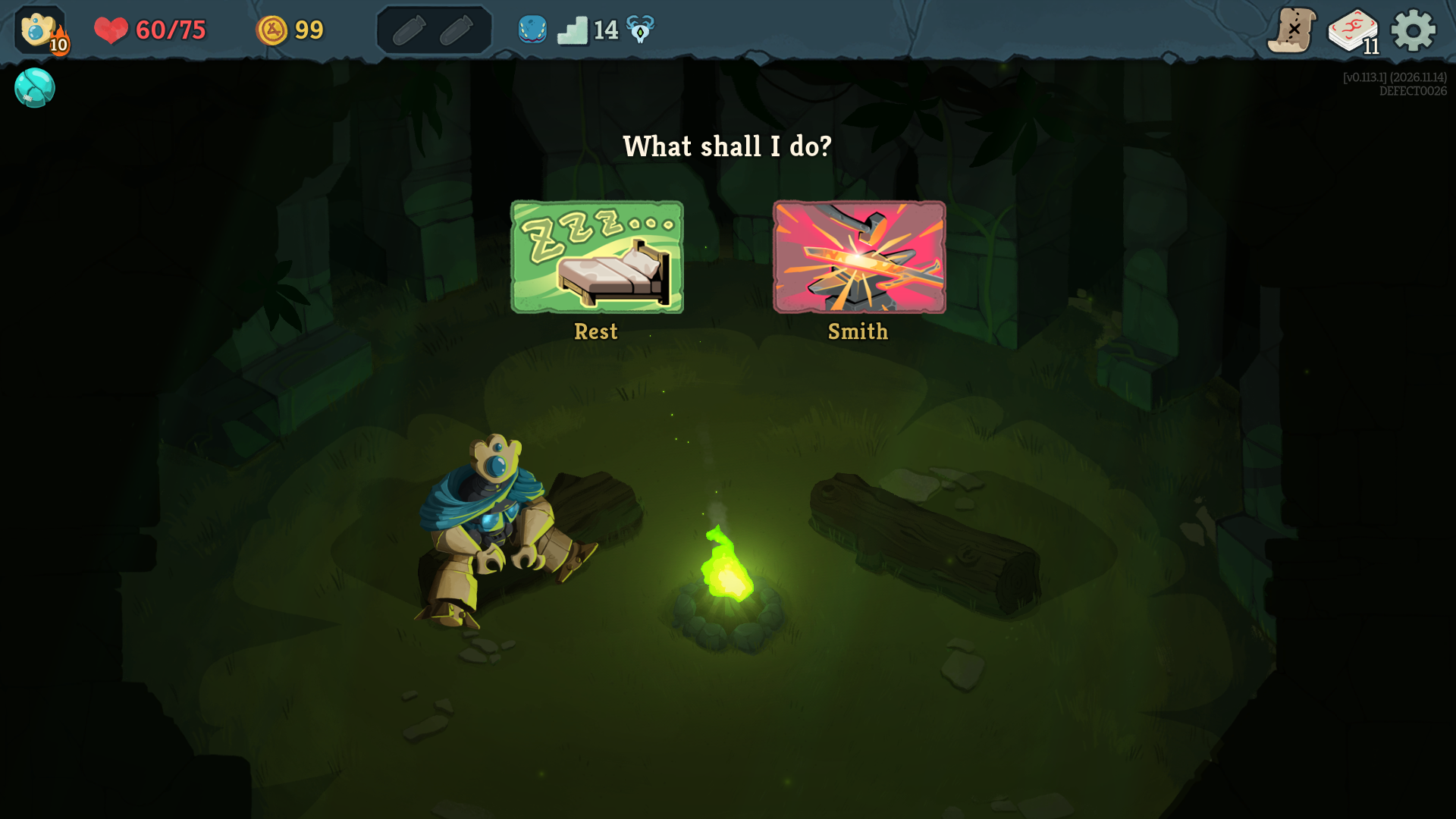}\hfill
  \includegraphics[width=0.49\linewidth]{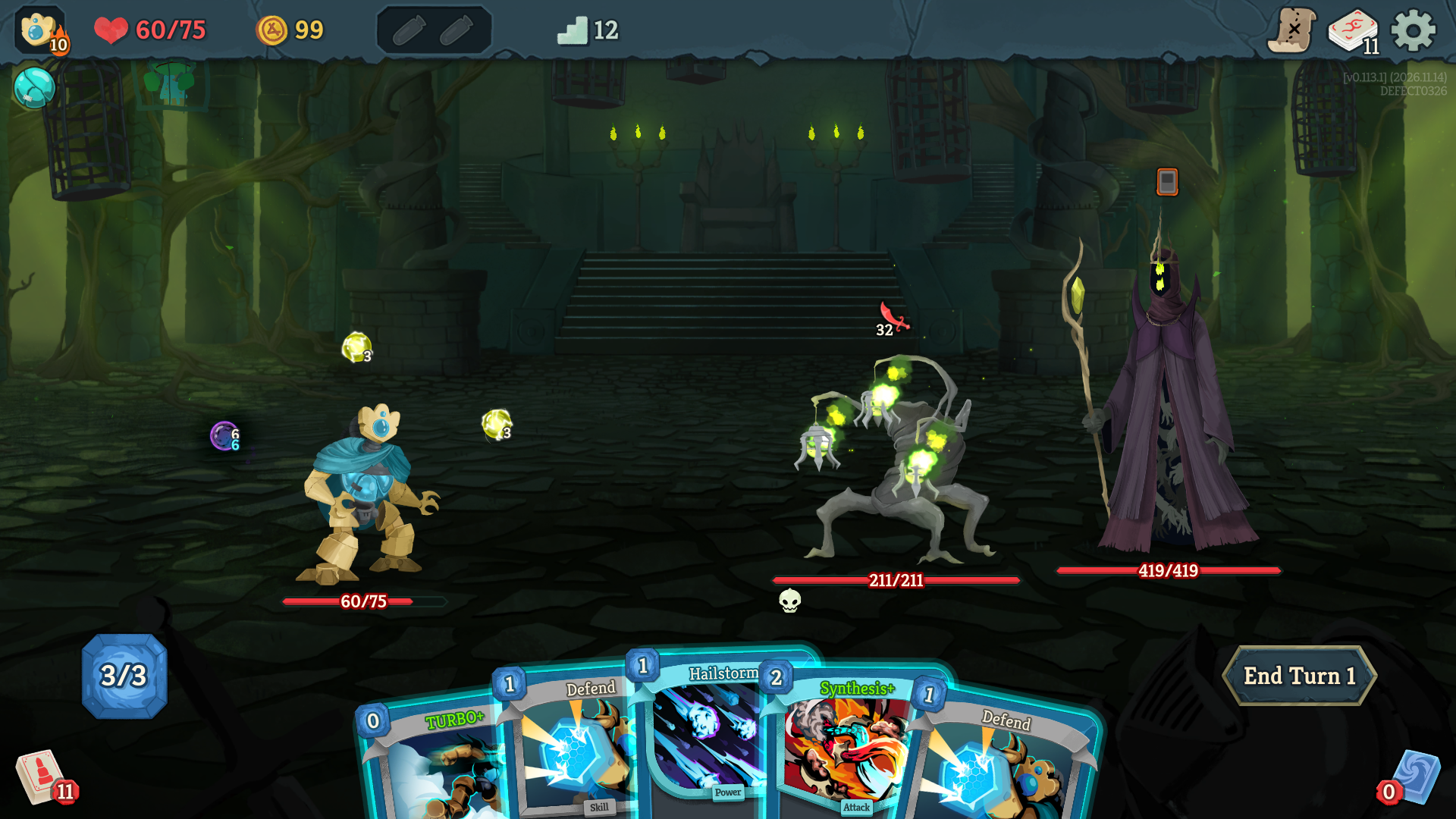}
  \caption{Screenshots of STS2 (v0.113.1) with the Defect at Ascension~10. Left: a non-combat decision at a rest site, where the player either rests to recover HP or upgrades a card (Smith). Right: a combat decision in a boss fight, with the hand of cards at the bottom, the remaining energy on the left, the Defect's orbs around it and the intents of the enemies above them. The player can choose cards to play from their hand each turn. The screenshots show Slay the Spire~2 by Mega Crit.}
  \label{fig:screenshots}
\end{figure}

STS2 is a video game released by Mega Crit.\footnote{As of November 2026, it is in early access.}
Figure~\ref{fig:screenshots} shows a decision outside combat and a decision in combat.
In each playthrough, called a run, the player starts with a small, weak deck and climbs through three acts.
Each act is a branching map of rooms, one per floor, that ends in a boss fight; the player chooses a path through normal fights, elite fights, events, shops, rest sites and treasure rooms.
A run that reaches the end spans about 50 floors.

Fights are turn-based.
Each turn the player draws cards and spends a limited amount of energy to play them, and may also use potions, which have one-time effects.
Attacks deal damage, skills include cards that grant block against incoming attacks, and powers have lasting effects.
Then the enemies act.
Each enemy shows its intent for the coming turn, so the player can plan around an attack; its later moves follow fixed patterns, except where its behavior branches at random.
Between fights the player builds the deck: after most fights the player may add one of several offered cards or skip them; at shops they can remove cards, at rest sites upgrade them, and at events trade hit points (HP), gold or cards for other benefits.
Relics, which give permanent effects, come from elites, shops, treasure rooms and events, including the Ancient that opens each act.

Each character has its own starting deck, starting relic and card pool, built around distinct mechanics: the Defect channels orbs, for example, and the Silent relies on poison, shivs and discards.
We evaluate on two public beta versions of the early-access game: v0.111.0, with five characters (the Ironclad, the Silent, the Defect, the Necrobinder and the Regent), and v0.113.1, which also includes the Fool, a sixth character introduced in v0.113.0.
Difficulty is set by the Ascension level, which stacks handicaps from level~1 to level~10, the highest level in these versions.
At Ascension~10 the third act ends with two boss fights in a row, the Act-3 bosses, and the run is won only if the player defeats both.
The Dread mode, added in v0.113.0, adds further handicaps to Ascension~10; unless stated otherwise, results in this paper are for the standard mode.

The player sees the cards in hand, the contents of the draw pile but not their order, the board, the map and the rules of the game.
Hidden are the order of the draw pile, the upcoming encounters and rewards, the random branches of enemy behavior and the state of the random number generators.
With a fixed seed, the game is deterministic given the player's actions.

Runs are long: a Defect run takes about 800 decisions, about 85\% of them in combat, and the legal actions change from one decision to the next with the cards in hand, the enemies on the board and the options on screen.

\subsection{Save and Load (SL), and no-SL}

The game saves a run automatically, and the (human) player can quit and continue it later.
In the original game, quitting in the middle of a fight and continuing returns the player to the start of that fight with the same random outcomes, so the player learns, for example, which cards will be drawn in the first turns.
Repeating this is called save and load (SL).

In AI research on Slay the Spire, SL refers to a more permissive setting in which the agent can access the random seed of the game, and can therefore make decisions with knowledge of future outcomes in advance.
In contrast, the no-SL setting restricts the agent to the information available to a human player, without access to the game's random seed.
Unless stated explicitly, all results in this paper are in the no-SL setting: the network sees only what a human player sees, and the test-time search (Section~\ref{sec:search}) resamples the hidden randomness instead of reading it.

\section{Method}

\subsection{The Learning Problem}
\label{sec:learning-problem}
We treat a single run of STS2, from the first floor until the player dies or defeats the final boss, as one episode of a partially observable Markov decision process.
Let $s_t$ denote the full internal game state at decision step $t$.
This includes information that is hidden from the player, such as the order of the draw pile, the upcoming encounters, and the state of the random number generators.
The agent instead receives an observation $o_t$, which contains everything visible to a human player, and a finite set of legal actions $\mathcal{A}(o_t)$ supplied by the game.
Depending on the decision type $b_t\in\{\mathrm{combat},\mathrm{non\text{-}combat}\}$, an action $a_t\in\mathcal{A}(o_t)$ is either a combat action (playing a card on a target, using a potion, or ending the turn) or a run-level choice (picking a card reward, choosing an event option, buying in a shop, resting or upgrading at a rest site, or selecting the next node on the map).
Executing $a_t$ moves the game to $s_{t+1}$ according to its rules and random processes, and yields a reward $r_t$.

The policy $\pi_\theta(a\mid o_t)$ is a distribution over $\mathcal{A}(o_t)$, and a single policy with parameters $\theta$ makes every decision in the run.
Its goal is to maximize the expected return
\begin{equation}
  J(\theta)=\mathbb{E}_{s_0\sim\rho_0,\,\pi_\theta}\!\left[\sum_{t=0}^{T-1}\gamma^t r_t\right],
\end{equation}
where $\rho_0$ is the distribution of initial states induced by the game seed, $T$ is the (random, finite) length of the run, and we use $\gamma=1$.
The reward $r_t$ is mainly given for defeating bosses and winning the run, so $J(\theta)$ is closely tied to the win rate.
For learning, we also use a value function
\begin{equation}
  V_\theta(o_t,z_t)\approx\mathbb{E}_{\pi_\theta}\!\left[\sum_{t'=t}^{T-1}\gamma^{t'-t}r_{t'}\,\middle|\,o_t,z_t\right],
\end{equation}
where $z_t$ is a part of the hidden state $s_t$, such as the actual draw order and the upcoming encounters.
The privileged input $z_t$ is used only during training; the test-time search (Section~\ref{sec:search}) queries the value head with $z_t$ set to zero.
We write $V_t=V_\theta(o_t,z_t)$, where the value is read from the value head for the decision type $b_t$.

\paragraph{Combat and non-combat decisions.}
A run interleaves two kinds of decisions that act on very different timescales.
A decision is non-combat ($b_t=\mathrm{non\text{-}combat}$) when it is taken outside a fight: map navigation, card rewards, events, shops, rest sites and treasure; every decision taken inside a fight, including choices that cards offer during the fight, is a combat decision.
A fight takes tens to hundreds of combat decisions, whereas the effect of a non-combat decision, such as adding a card to the deck, shows only through the fights that follow it.
We therefore use one policy network with separate scoring and value heads for the two decision types, and train them with credit assigned on two different timescales (Section~\ref{sec:training}).
Seen only at its non-combat decisions, a run is a semi-Markov decision process~\cite{bradtke1995smdp,sutton1999options}: each non-combat decision is followed by a variable number of combat decisions before the next one, and its consequence is the sum of all rewards obtained in between. 
Because the two kinds of decisions share one encoder, each can benefit from what the other learns: for example, what the network learns about a card's effect in combat can inform whether to take that card as a reward. We expect this sharing to help.

\subsection{Observation Encoding and Network}
\label{sec:observation}

The observation gives the identity of each card, relic, enemy and other piece of content, through a learned embedding of its vocabulary ID and a learned projection of its row in a content table, and adds consequences of public information computed by the engine; Section~\ref{sec:ablation} removes each of the two in turn.

\paragraph{Entity records.}
Every decision is a choice among the legal candidates listed by the engine, the process that runs the game headlessly (Section~\ref{sec:training}).
The observation consists of entity records (the player, an enemy, a card, relic, potion, power, map node, event option and so on), each with integer fields for its kind, vocabulary ID, zone, owner and an auxiliary ID, and a vector of numerical features, and of candidates, each referring to up to two records, such as a card and its target (Figure~\ref{fig:system}b).
Each vocabulary ID is also described by a fixed row of a content table extracted from the game data (cost, type, rarity, keywords, numerical variables, effect signature), whose learned projection is added to the ID embedding, so that new content starts near similar content.

\paragraph{Consequences computed by the engine.}
The engine also supplies consequences of public information that a network would otherwise have to compute, counting only effects whose outcome is already determined: the damage and block of each card after all modifiers, the HP the player would lose by ending the turn now, and the HP that would remain after a given action followed by ending the turn.
The draw pile is presented sorted by card, so its order is not revealed, and enemy moves beyond the current intent are given as probabilities taken from the enemies' move patterns, never from the random number generator.

\paragraph{Network.}
Powers and enemy intents are added to the records of the creatures they belong to, and the draw pile, discard pile, exhaust pile, deck and relics are each summarized by a mean-pooled slot with a count term.
A six-layer pre-norm Transformer (width 256, 4 heads) encodes the remaining records together with these slots.
A multilayer perceptron (MLP) scores each candidate from the encoded records it refers to, its own features and a global vector; a softmax over the candidates gives the policy.
Combat and non-combat decisions have separate scoring and value heads.
Following asymmetric actor--critic learning~\cite{pinto2017asymmetric}, only the value heads and an auxiliary head that predicts the outcome of the current fight receive privileged inputs: the actual draw order, the pre-rolled upcoming encounters, the next random enemy move and the pre-generated content of the run.
They are produced only for training: evaluation, play in the released game and search do not request them (Suphx~\cite{li2020suphx}, by contrast, gives hidden information to the policy and removes it gradually).
A critic that conditions on part of the hidden state can bias the advantage estimates of a policy that does not see it~\cite{baisero2022unbiased}; we did not measure this bias, and search queries the value head with these inputs set to zero, a condition that a diagnostic in Appendix~\ref{sec:test-time-inference} examines.
A future-readout head predicts, for each combat candidate, what follows over the next 12 decisions and adds a learned correction, initialized to zero, to its score; the Ironclad's network also runs its encoder twice, feeding the policy of the first pass into the second.

\subsection{Training Procedure}
\label{sec:training}

\begin{figure}[t]
\centering
\begin{tikzpicture}[
    font=\footnotesize,
    >={Stealth[length=1.7mm,width=1.3mm]},
    box/.style={draw, rounded corners=1.5pt, fill=black!6, align=center, inner sep=2.5pt},
    dark/.style={box, fill=black!17},
    lbl/.style={font=\scriptsize, align=left, inner sep=1.5pt},
    ttl/.style={font=\footnotesize\bfseries, anchor=north west, inner sep=0pt},
  ]
  \node[ttl] at (0,0) {(a) Training system};
  \node[box, text width=28mm] (full) at (1.5,-1.25) {Full runs\\(half of the engines)};
  \node[box, text width=28mm] (prac) at (1.5,-2.25) {Practice episodes\\(the other half)};
  \node[box, text width=11mm, minimum height=18.5mm] (states) at (7.3,-1.75) {Saved\\states};
  \draw[->] (full.east) -- node[lbl, above] {state at each map decision} (full.east -| states.west);
  \draw[->] (prac.east -| states.west) -- node[lbl, above] {replay with fresh randomness} (prac.east);
  \node[draw, densely dotted, rounded corners=2pt, inner sep=3pt, fit=(full)(prac)(states)] (actors) {};
  \node[lbl, anchor=south west, inner sep=1pt] at (actors.north west) {Actors running engines (the headless game) on CPUs};
  \node[dark, text width=40mm] (learner) at (2.1,-4.1) {\textbf{Learner} (GPU)\\GAE on two chains, PPO, EMA};
  \node[box, text width=27mm] (eval) at (6.7,-4.1) {Routine evaluation\\(greedy, fixed seeds)};
  \draw[->] (0.8,0 |- actors.south) -- node[lbl, right] {run segments,\\practice episodes} (0.8,0 |- learner.north);
  \draw[<-] (3.3,0 |- actors.south) -- node[lbl, right] {weights after\\every update} (3.3,0 |- learner.north);
  \draw[->] (learner) -- node[lbl, above] {weights} (eval);
  \begin{scope}[xshift=8.5cm]
    \node[ttl] at (0,0) {(b) Network};
    \node[box, text width=22mm, anchor=north] (rec) at (1.2,-0.5) {Entity records:\\ID embedding\\+ content table\\+ features};
    \node[box, text width=15mm, anchor=north] (own) at (3.62,-0.5) {Powers and intents, added to owners};
    \node[box, text width=17mm, anchor=north] (pool) at (5.9,-0.5) {Piles, deck, relics:\\5 pooled slots};
    \node[dark, minimum width=68mm] (enc) at (3.42,-2.55) {Transformer encoder (6 layers, width 256)};
    \draw[->] (rec) -- (rec |- enc.north);
    \draw[->] (own) -- (own |- enc.north);
    \draw[->] (pool) -- (pool |- enc.north);
    \node[box, text width=31mm] (score) at (1.68,-3.8) {Candidate scorer\\combat $\mid$ non-combat\\(+ future readout)};
    \node[box, text width=28.5mm] (value) at (5.42,-3.8) {Value heads\\combat $\mid$ non-combat};
    \draw[->] (score.north |- enc.south) -- (score.north);
    \draw[->] (value.north |- enc.south) -- (value.north);
    \node[lbl, align=center, anchor=north] (pi) at ([yshift=-3.5mm]score.south) {policy: softmax over the\\legal candidates};
    \draw[->] (score.south) -- (pi.north);
    \node[draw, dashed, rounded corners=1.5pt, align=center, inner sep=2pt, font=\scriptsize, anchor=north] (priv)
         at ([yshift=-3.5mm]value.south) {privileged inputs\\(training only)};
    \draw[->, dashed] (priv.north) -- (value.south);
  \end{scope}
\end{tikzpicture}
\caption{(a) Practice episodes replay fights from states saved in full runs; full runs reach the learner in segments, practice episodes whole. (b) A candidate is scored from the encoded records it refers to, its own features and a global vector; only the value heads (and an auxiliary fight-outcome head, not shown) receive privileged inputs.}
\label{fig:system}
\end{figure}
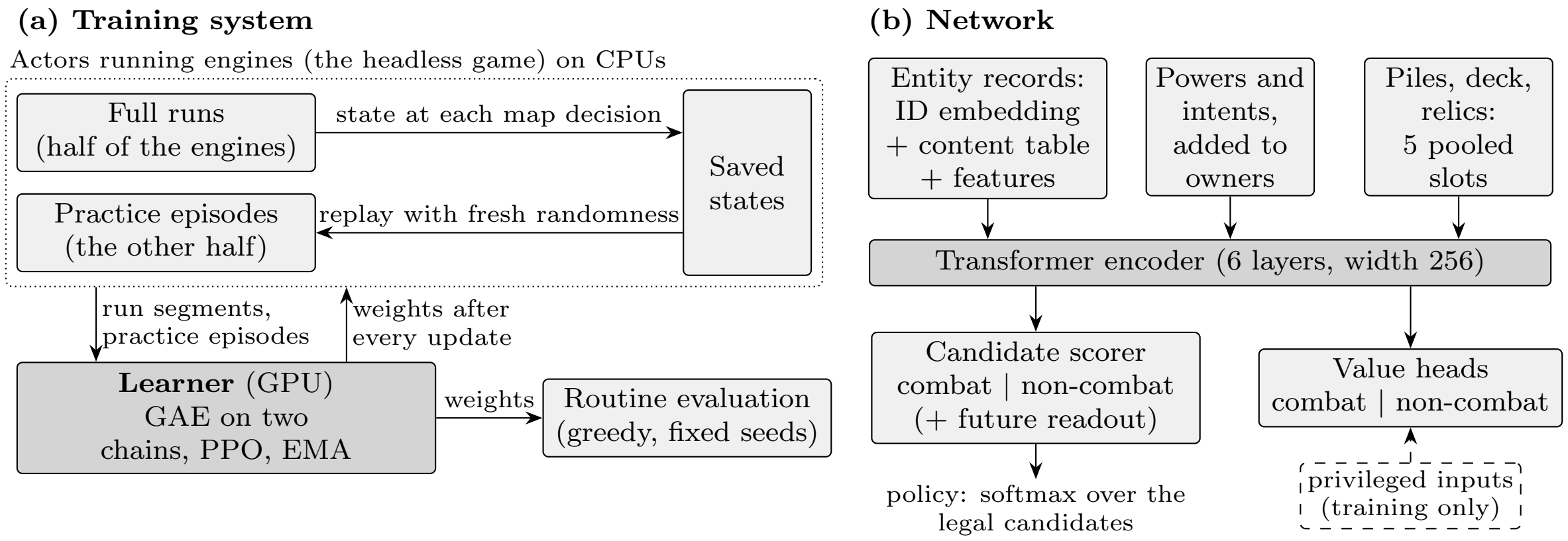

We describe the recipe in its final form, as used for the Fool, the Ironclad and the Necrobinder (Figure~\ref{fig:system}, Algorithm~\ref{alg:update}); earlier characters differed in a few components (Table~\ref{tab:recipe}), and Appendix~\ref{sec:app-recipe} gives all hyperparameters and how the recipe developed.

\paragraph{Environment.}
The agent is trained and evaluated on the game's own rules.
A headless host loads the game's compiled assembly into a .NET~9 process and replaces only the game's presentation layer, so the training environment is the game itself rather than a reimplementation; we call one such process an engine.
The host runs all asynchronous game logic on one thread, so a seed and a sequence of actions reproduce a run exactly.
It can also save the player's state (deck, relics, potions, HP and gold) in the game's own save format and later start a practice fight from it, with new random outcomes.
Each character is trained with one learner GPU at a time (two in the Defect's first days) and a few hundred engines on the CPUs of one or two machines (Table~\ref{tab:recipe}); with batched inference, this setup collects about 5,000--10,000 decisions per second.
In all, the six evaluated networks took 25.0 days and 14.2 billion decisions of training (Tables~\ref{tab:recipe} and~\ref{tab:main-results}).
These figures cover only the training runs that produced the evaluated networks, including the method development that happened within the Defect's run; they do not count distillation, evaluation, search experiments, or runs and branches that led to no evaluated network, so they are a lower bound on the compute of the project.

\paragraph{Full runs and combat practice.}
Half of the engines play full runs, and the rest play \emph{combat practice}: practice episodes, each a single fight or the two Act-3 bosses in sequence, that concentrate experience on the fights in which runs are lost.
At every map decision of a full run, the engine saves the player's state.
A practice episode normally replays an upcoming fight from such a state with fresh (new) randomness, so the policy practices with the deck, relics, potions and HP that it accumulated itself.
This is related to curricula that start episodes from states near a goal or along a demonstration~\cite{florensa2017reverse,salimans2018montezuma} and to returning to saved states, as in Go-Explore~\cite{ecoffet2021goexplore}; here the states are the agent's own and what follows them is drawn anew.
Combat practice covers the boss of each act, the normal and elite fights of each act, the two Act-3 bosses played in sequence, and fights with randomly generated decks.
Act-3 boss pairs are sampled in proportion to $[\hat p(1-\hat p)]^3$, where $\hat p$ is a win probability derived from the value head (details in Appendix~\ref{sec:app-recipe}); this favors practice episodes with an uncertain outcome, in the spirit of sampling by learning potential in Prioritized Level Replay~\cite{jiang2021plr}.
Second-boss positions are weighted by their recent loss rate, and the other categories are drawn uniformly from recent states.
The mix is shifted toward the Act-3 bosses as training progresses; at the end, boss fights make up 52--55\% of practice episodes, half or more of them in Act~3.
Actors, the processes that run engines, send full runs to the learner in segments, cut at the first map decision after at least 64 new decisions.

\paragraph{Rewards.}
Full runs are rewarded for progress: $+0.02$ per floor, $+0.1$ for the Act-1 boss ($+0.5$ for the Necrobinder), $+0.5$ for the Act-2 boss, $+1.5$ for the first Act-3 boss and $+4$ for winning the run, that is, beating the second Act-3 boss; a won run thus earns about 7 in total.
For the Defect, from update 164,606 the rewards before the win were withheld and paid in full at a win, so that only lost runs earned less (Table~\ref{tab:recipe}).
The only other term is a penalty of 0.1 for each potion discarded without taking a new one before the next map decision; our engine offers discarding only when a slot is needed for a new potion.
There is no shaping for HP, relics or upgrades.
A practice episode is rewarded $+1$ for a win and $-1$ for a loss, plus the change in HP as a fraction of maximum HP, with a bonus for won boss fights and partial credit for enemy HP removed in lost ones; an Act-3 pair is scored on both fights together, without the HP term.

\begin{algorithm}[t]
\caption{Training of AlphaDefect. The actors and the learner run concurrently.}
\label{alg:update}
\small
\begin{algorithmic}[1]
\Statex \textbf{Actors} (run continuously)
\Loop
  \State Play a full run, saving the state at each map decision in a store $\mathcal{S}$, or a practice episode from a state in $\mathcal{S}$ with fresh randomness
  \State Send the decisions to the learner (full runs in segments)
\EndLoop
\Statex \textbf{Learner} (one update)
\State Collect a batch of $B$ recent decisions
\State Compute $\hat A_t$ for combat decisions (Eq.~\ref{eq:gae}) and $\hat A^{\mathrm{n}}_k$ for non-combat decisions (Eqs.~\ref{eq:macro-return}--\ref{eq:macro-adv})
\State Standardize the two groups of advantages separately
\For{two epochs over minibatches}
  \State Take a PPO step (Muon for matrix parameters, Adam for the rest)
\EndFor
\State Update the EMA weights $\bar\theta \gets 0.999\,\bar\theta + 0.001\,\theta$; publish $\theta$ to the actors
\end{algorithmic}
\end{algorithm}

\paragraph{Policy optimization.}
We train with PPO~\cite{schulman2017ppo} (clipping parameter 0.2, two epochs per update, value-loss coefficient 0.5) with entropy coefficients of 0.01 for combat and 0.05 for non-combat decisions.
An update is one PPO iteration on a batch of recently collected decisions (Algorithm~\ref{alg:update}); we identify checkpoints by the number of updates before them.
Credit is assigned on two timescales, following the separation of combat and non-combat decisions in Section~\ref{sec:learning-problem}.
For combat decisions, we use generalized advantage estimation (GAE)~\cite{schulman2015gae} along the sequence of all decisions,
\begin{equation}
  \delta_t=r_t+\gamma V_{t+1}-V_t,\qquad
  \hat A_t=\sum_{l\ge 0}(\gamma\lambda)^l\,\delta_{t+l},
  \label{eq:gae}
\end{equation}
with $\lambda=0.95$ and $V_T=0$.
The sum is cut, and bootstrapped from the value there, at the end of a segment and at the next map decision after each fight (the choice of the next room), or at the next fight if that comes first.

For non-combat decisions, we compute GAE along the sequence of non-combat decisions only, treating the run as a semi-Markov decision process over them~\cite{bradtke1995smdp,sutton1999options}.
Let $\tau_0<\tau_1<\cdots<\tau_K$ be the steps at which the non-combat decisions of a run are taken.
All rewards between two non-combat decisions, including those of any intervening fight, are credited to the earlier one, and after the last one the sum extends to the end of the run:
\begin{equation}
  R_k=\sum_{t=\tau_k}^{\tau_{k+1}-1}r_t\quad(k<K),\qquad
  R_K=\sum_{t=\tau_K}^{T-1}r_t.
  \label{eq:macro-return}
\end{equation}
The advantage of the $k$-th non-combat decision (superscript $\mathrm{n}$ for non-combat) is
\begin{equation}
  \delta^{\mathrm{n}}_k=R_k+V_{\tau_{k+1}}-V_{\tau_k},\qquad
  \hat A^{\mathrm{n}}_k=\delta^{\mathrm{n}}_k+\lambda_{\mathrm{n}}\hat A^{\mathrm{n}}_{k+1},
  \label{eq:macro-adv}
\end{equation}
with $V_{\tau_{K+1}}=0$, $\hat A^{\mathrm{n}}_{K+1}=0$ and $\lambda_{\mathrm{n}}=0.95$, and it replaces $\hat A_{\tau_k}$ at the non-combat decisions.
A card pick is thus judged by the full outcome of the fights that follow it, and its credit decays by $\lambda_{\mathrm{n}}$ once per later non-combat decision rather than once per combat decision.
The value targets are $\hat A_t+V_t$ for combat and $\hat A^{\mathrm{n}}_k+V_{\tau_k}$ for non-combat decisions, and in each update the advantages of the two groups are standardized separately before they enter the PPO loss,
\begin{equation}
  L(\theta)=-\hat{\mathbb{E}}_t\Bigl[\min\bigl(\rho_t\tilde A_t,\ \operatorname{clip}(\rho_t,1-\epsilon,1+\epsilon)\,\tilde A_t\bigr)\Bigr]
  +c_V\,\hat{\mathbb{E}}_t\bigl[(V_\theta(o_t,z_t)-\hat R_t)^2\bigr]
  -\hat{\mathbb{E}}_t\bigl[c_{b_t}\,\mathcal{H}\bigl(\pi_\theta(\cdot\mid o_t)\bigr)\bigr]
  +L_{\mathrm{aux}},
  \label{eq:ppo-loss}
\end{equation}
where $\hat{\mathbb{E}}_t$ averages over a minibatch, $\rho_t=\pi_\theta(a_t\mid o_t)/\pi_{\theta_{\mathrm{old}}}(a_t\mid o_t)$ with $\pi_{\theta_{\mathrm{old}}}$ the policy that produced the decision, $\tilde A_t$ is the standardized advantage of step $t$ ($\hat A_t$ for combat and $\hat A^{\mathrm{n}}_k$ for non-combat decisions, $t=\tau_k$), $\hat R_t$ is the value target, computed with the values $V_t$ recorded when the decision was made, $\epsilon=0.2$, $c_V=0.5$, $c_{b_t}$ is 0.01 for combat and 0.05 for non-combat decisions, and $L_{\mathrm{aux}}$ collects the losses of the categorical value head and the auxiliary heads (Appendix~\ref{sec:app-recipe}).
Similar two-level credit assignment has been combined with PPO in hierarchical reinforcement learning with options, where a higher-level decision is credited with the outcome of the segment it starts~\cite{peng2026hiper}, and option-level and action-level policies are trained with PPO~\cite{klissarov2017ppoc}, in one implementation with separately normalized advantages~\cite{cannon2025fracos}; in our case, the switches between the two levels follow from the rules of the game (a fight starts when the player enters a combat room and ends when one side is defeated) rather than from a learned termination function.

The chain of non-combat decisions is cut at segment ends as well: Eqs.~\ref{eq:macro-return}--\ref{eq:macro-adv} are applied within each segment, with $\tau_K$ the last non-combat decision of the segment and $T$ the first step after the segment; the value of the first decision of the next segment, a map decision, takes the place of $V_{\tau_{K+1}}$, which is 0 only at the end of the run, and $\hat A^{\mathrm{n}}_{K+1}=0$ at every segment end.
The learner discards decisions produced by weights more than six updates old, except in pieces (a run segment or a practice episode) of at least 1,000 decisions.

\paragraph{Optimizer, schedule and weight averaging.}
Muon~\cite{jordan2024muon} updates the hidden weight matrices, about 43\% of the parameters, and Adam~\cite{kingma2015adam} the rest.
Routine evaluation plays the network greedily on a fixed pool of 192 to 1,000 seeds, separate from those of training games and from the test seeds.
When the win rates in routine evaluation and in combat practice stop rising for about a thousand updates, we change the learning rates and the batch size by hand: we halve the learning rates and double the batch size $B$ (up to 131,072 decisions), and later accumulate gradients or halve the learning rates alone.
This rule follows Beukman et al.~\cite{beukman2026stagnation}, who trace the plateaus of PPO to policy updates that are too large for the noise in their sample-based estimates and reduce that noise by taking smaller steps or collecting more samples per update; halving the learning rate and doubling the batch does both.
In the training of the Regent and the Silent, the largest single gains followed these changes (Table~\ref{tab:evidence}).
We also keep an exponential moving average (EMA) of the weights with decay 0.999 per update, a form of iterate averaging~\cite{polyak1992averaging}; the EMA weights are the evaluated network for the Silent, the Fool and the Ironclad.
For the Defect and the Regent, the evaluated network is a student distilled~\cite{hinton2015distilling,rusu2016distillation} from an ensemble of checkpoints of the same training run, related to snapshot ensembles~\cite{huang2017snapshot} (Appendix~\ref{sec:app-recipe}).

\paragraph{Differences between characters and the evidence for each component.}
The recipe evolved over the characters, trained in the order of Table~\ref{tab:recipe} (the last two in parallel), which lists where they differ.
Early in its training, the Defect also used reward shaping, sent whole runs instead of segments and, for about two days, drew half of its practice episodes against bosses from saved states that a perfect-information fight solver, CombatSolver~\cite{combatsolver}, had found winnable (Appendix~\ref{sec:app-recipe}); one of us (T.~Yu) is a main contributor to CombatSolver.
Table~\ref{tab:evidence} summarizes the evidence behind each component.
Test-time search, ensemble distillation and weight averaging were tested against a control on the same seeds, as were the two parts of the observation encoding and combat practice late in the Ironclad's training (Section~\ref{sec:ablation}); several components were never tested in isolation.
We report interval estimates, as Agarwal et al.~\cite{agarwal2021precipice} recommend for deep reinforcement learning; with one training run per character, they cover the sampling of seeds but not the variation between runs.

\begin{table}[t]
\centering
\footnotesize
\setlength{\tabcolsep}{3.0pt}
\caption{Differences in training between the six characters. Observation format: revision of the observation layout. Learning rates are those of the evaluated network (Adam, then Muon where used). Fight cut: the combat GAE is cut at the end of each fight (``map'': at the next map decision). GPUs are NVIDIA RTX models, in order of use; engines are those playing training games at the end of training, with the CPU threads of each machine. Days of training: wall-clock days from the first update to the evaluated network (for distilled networks, to their starting checkpoint), pauses included.}
\label{tab:recipe}
\smallskip
\begin{tabular}{>{\raggedright\arraybackslash}p{27mm}cccccc}
\toprule
 & Defect & Regent & Silent & Fool & Ironclad & Necrobinder \\
\midrule
Game versions & v0.107.1, v0.111.0 & v0.111.0 & v0.111.0 & v0.113.0, v0.113.1 & v0.113.1 & v0.113.1 \\
Observation format & 19 & 20 & 23 & 27 & 28 & 28 \\
Optimizer & Adam & Adam & Adam, Muon & Adam, Muon$^{a}$ & Adam, Muon & Adam, Muon \\
Learning rate ($10^{-5}$) & 5.5 & 20 & 10, 15 & 15, 7.5 & 3.75, 7.5 & 7.5, 15 \\
Act-1 boss reward & 0.5 & 0.5 & 0.1 & 0.1 & 0.1 & 0.5 \\
Progress rewards & at a win$^{c}$ & in the run & in the run & in the run & in the run & in the run \\
Engines in full runs & 3/4 & 1/2 & 1/2 & 1/2 & 1/2 & 1/2 \\
Fight cut & no & yes & yes & map & map & map \\
Solver-filtered combat practice & early & -- & -- & -- & -- & -- \\
Skipping certain fights & -- & -- & late & -- & -- & -- \\
Future-readout head & -- & -- & -- & yes & yes & yes \\
Encoder run twice & -- & -- & -- & -- & yes & -- \\
Evaluated network & distilled & distilled & EMA & EMA & EMA & raw \\
Learner GPU (RTX) & 3090, 4090$^{b}$ & 4090 & 4090 & 4090 & 3090, 4090$^{b}$ & 4090 \\
Engines & 336 + 160 & 336 + 160 & 336 & 336 & 162, 336$^{b}$ & 336 \\
CPU threads & 32 + 36 & 32 + 36 & 32 & 32 & 32 & 32 \\
Days of training & 10.9 & 2.7 & 2.7 & 3.5 & 3.0 & 2.2 \\
\bottomrule
\end{tabular}

\smallskip
\parbox{0.92\linewidth}{\footnotesize $^{a}$Schedule-free Adam, with schedule-free averaging also applied to Muon.\\
$^{b}$The Defect's learner used both GPUs in its first days and the RTX~4090 from update 50,181; the Ironclad's moved to the RTX~4090, with 336 engines, at update 24,280.\\
$^{c}$From update 164,606 (over 300 updates), the rewards per floor and for the bosses before the last one were withheld during the run and paid in full at a win (Appendix~\ref{sec:app-recipe}).}
\end{table}

\begin{table}[tp]
\centering
\small
\caption{Evidence for the components of the recipe. Paired: two networks or settings compared on the same test seeds. Before and after: readings before and after a change during training, which also include the effect of further training.}
\label{tab:evidence}
\smallskip
\begin{tabular}{>{\raggedright\arraybackslash}p{0.28\linewidth}>{\raggedright\arraybackslash}p{0.22\linewidth}>{\raggedright\arraybackslash}p{0.415\linewidth}}
\toprule
Component & Evidence & Result \\
\midrule
Entity identities removed at the input & paired, 1,000 seeds (Ironclad, continued from 0.60 billion decisions) & no run won, against 57.3\% for the control at 0.74 billion decisions (Section~\ref{sec:ablation}) \\
Engine-computed consequences removed & as above & $-8.3$ points at 0.74 billion decisions ($p=2\times10^{-5}$), $-4.9$ at 0.68 ($p=0.015$) \\
Combat practice removed & as above & $+0.8$ points at 0.74 billion decisions ($p=0.72$), $+2.9$ at 0.68 ($p=0.16$) \\
Test-time search & paired, 1,000--3,000 seeds & $+2.7$ to $+5.8$ points (Table~\ref{tab:main-results}); $+2.4$ to $+5.6$ on fresh seeds (Table~\ref{tab:retest}) \\
Ensemble distillation & paired, 1,000--3,500 seeds & Defect: a student with the architecture of its starting checkpoint $+3.5$ points over it ($+4.0$ on the seeds not used to choose it); the evaluated student, which also dropped one attention component, $+6.4$ over its own starting checkpoint on fresh seeds ($p=2\times10^{-4}$) and $+5.4$ on the seeds of Table~\ref{tab:main-results}; Regent: the evaluated student $-0.9$ against the EMA network it started from on fresh seeds ($p=0.61$) and $+1.7$ on the seeds of Table~\ref{tab:main-results} ($p=0.33$), although its teacher ensemble was $+5.2$ over a single checkpoint on the seeds used to choose it; Silent: ensembles no better than its EMA network \\
Weight averaging (EMA) & paired, 500--1,000 seeds & Regent $+4.4$ over a single checkpoint ($p=0.013$); Silent $+4.8$ over the raw weights of the same version at one checkpoint ($p=0.006$) and $+0.6$ at an earlier one; Ironclad $+2.4$ ($p=0.18$); Defect at its plateau $-0.4$ ($p=0.93$) \\
Sending runs in segments & before and after (Defect) & 12.6\% to 29.4\% on the same 1,000 seeds \\
Halving the learning rate and doubling the batch & before and after & Regent 46\% to 56\%; Silent 29\% to 47\% and 53\% to 63\% (routine evaluation) \\
Skipping certain fights & before and after (Silent) & a third more full runs per hour; no measured change in strength \\
Muon, switched on in the middle of the Regent's training & before and after (Regent) & full-run win rate in training fell by about 5 points; rolled back \\
Card and relic embeddings zeroed at load (Regent, released game) & accidental; not paired, different seeds & 11 of 28 runs won (39.3\%, 95\% Wilson interval 23.6--57.6) against 117 of 151 (77.5\%, 70.2--83.4) with the corrected checkpoint of the same version (Section~\ref{sec:live-results}) \\
Other components & not tested in isolation & solver filter, two credit timescales, privileged value inputs, the content table apart from the ID embeddings, Muon from the start, future-readout head; the observation encoding and combat practice from the start of training \\
\bottomrule
\end{tabular}
\end{table}

\subsection{Test-Time Search}
\label{sec:search}

The trained network can also combine with a lookahead search. 
We only used the search at decisions with more than one candidate where the most probable candidate has probability below 0.9; all other decisions, including every decision outside combat, are played greedily based on the trained network.
It is designed to use only public information: each simulation starts from the current position, resamples the hidden state (the order of the draw pile and the state of the random number generators; Appendix~\ref{sec:test-time-inference} lists the known exceptions), and evaluates positions with the value head, whose privileged inputs are set to zero.

For the characters of v0.111.0, a rollout search over sampled worlds runs in the two Act-3 boss fights.
It compares up to four candidates with at most 32 rollouts in two rounds of sequential halving, each round in sampled worlds shared by the candidates it compares: in each rollout, the candidate is played, the network continues greedily for 12 decisions, and the value head scores the result; the search departs from the greedy candidate only if the paired difference in value exceeds two standard errors.
These rollouts run in sts2-sim~\cite{sts2sim}, a reimplementation of the rules that copies combat states quickly, while the moves are played in the game; a copy of the position in sts2-sim follows the game, and where the two disagree the search falls back to greedy play (at 1.3--1.4\% of the decisions at which a search was due, in the Defect and Regent evaluations).
For the characters evaluated on v0.113.1, an information-set tree search runs in all boss and elite fights: a closed-loop tree search with 32 simulations that uses the network's policy as the prior and its value head at the leaves, without rollouts.
Both searches improve a fixed policy at decision time and keep its choice unless the search finds a better one, as in search over a fixed blueprint policy in partially observable games~\cite{lerer2020sparta}.
Appendix~\ref{sec:test-time-inference} describes both methods, the one leak of hidden information that we found and fixed, the residual leaks, and the variants that did not help.

\section{Results}
\label{sec:results}

We report win rates at Ascension~10 for six characters: the Defect, the Regent and the Silent on v0.111.0, and the Fool, the Ironclad and the Necrobinder on v0.113.1.
For each character we evaluate one network, first with greedy play and then, where available, with the test-time search of Section~\ref{sec:search}.
We then report runs in the released game, analyze where runs are lost, and compare the results with community statistics and with other agents.

\subsection{Evaluation Protocol}
\label{sec:eval-protocol}

An evaluation plays a fixed network on a fixed list of seeds in the headless engine.
Because the engine is deterministic (Section~\ref{sec:training}), two policies evaluated on the same seeds start from the same conditions, and we compare them seed by seed.
We give 95\% Wilson intervals for win rates~\cite{wilson1927}.
For a paired comparison we report the difference in win rate with a 95\% Wald interval for paired proportions and an exact McNemar test~\cite{mcnemar1947} on the seeds won by only one of the two policies.
All intervals in this paper are 95\% intervals.
They reflect the sampling of seeds for one fixed network and do not include the variation between checkpoints of the same training run, which for the Defect at its plateau was about 1.2 percentage points in standard deviation on 1,000 seeds.

On v0.113.1, evaluations apply two rules that end games that would otherwise not finish.
First, when the same position occurs for the fourth time in a row within one player turn while the enemies' HP and block barely change, the engine ends the turn.
Second, a fight is lost after 20 consecutive player turns in which the enemies' total HP reaches no new low and no enemy dies.
The v0.111.0 evaluations in Table~\ref{tab:main-results} were run without these rules; on the Silent's 1,000 seeds, applying them changed the outcome of one run, and on the fresh seeds of Table~\ref{tab:retest}, of two runs for the Silent and three for the Regent.

Test seeds come from pools drawn separately from the seeds used for routine evaluation during training, and games played during training use seeds of a different form, so no test seed is played during training.
Some test seeds were nevertheless used to choose between networks or settings, marked with $\dagger$ in Table~\ref{tab:main-results}.
The Regent's distillation teacher was chosen among ensembles evaluated on its test seeds, and the Silent's network among checkpoints evaluated on its; for the Fool, both the network and the search setting were chosen on its test seeds.
The Ironclad's network was chosen on another pool, and its row uses seeds that were not used for any choice.
Section~\ref{sec:main-results} repeats these evaluations on seeds drawn after all choices had been made (Table~\ref{tab:retest}).
The Defect's 1,500 seeds were not used to choose its network; they were later used to compare it with three challengers, none of which replaced it.
The Necrobinder's network, the checkpoint at the end of its training, was not chosen on any seeds; its row uses the first 1,000 seeds of the pool of Table~\ref{tab:retest}, drawn after all choices had been made.

\subsection{Win Rates at Ascension~10}
\label{sec:main-results}

\begin{table}[t]
\centering
\small
\caption{Win rates (\%) at Ascension~10 in the headless engine, with 95\% Wilson intervals in parentheses.
Training: decisions used to train the evaluated network, in billions, practice episodes included (for distilled networks, up to their starting checkpoint).
Seeds: test seeds per evaluation.
Greedy: the most probable candidate at every decision.
Search: test-time search (Section~\ref{sec:search}) in the fights named in each group; all other decisions are greedy.
$\Delta$: difference between search and greedy on the same seeds, in percentage points, with 95\% interval.
Network: \emph{distilled}, a single network trained to match the averaged policy of several earlier checkpoints of the same training run; \emph{EMA}, an exponential moving average of the training weights; \emph{raw}, the training weights themselves.}
\label{tab:main-results}
\smallskip
\begin{tabular}{llrrccc}
\toprule
Character & Network & Training & Seeds & Greedy & Search & $\Delta$ \\
\midrule
\multicolumn{7}{l}{\emph{v0.111.0; rollout search in the two Act-3 boss fights}} \\
Defect   & distilled & 6.7 & 1,500 & 76.7 (74.5--78.8) & 80.3 (78.2--82.3) & $+3.6$ (2.1--5.1) \\
Regent   & distilled & 2.1 & 1,000\rlap{$^{\dagger}$} & 74.1 (71.3--76.7) & 77.2 (74.5--79.7) & $+3.1$ (1.2--5.0) \\
Silent   & EMA       & 1.3 & 1,000\rlap{$^{\dagger}$} & 71.3 (68.4--74.0) & 74.0 (71.2--76.6) & $+2.7$ (1.1--4.3) \\
\midrule
\multicolumn{7}{l}{\emph{v0.113.1; information-set tree search in all boss and elite fights}} \\
Fool     & EMA       & 1.3 & 3,000\rlap{$^{\dagger}$} & 62.2 (60.5--63.9) & 68.0 (66.3--69.6)$^{a}$ & $+5.8$ (4.1--7.5) \\
Ironclad$^{b}$ & EMA & 1.4 & 1,000 & 67.9 (64.9--70.7) & 73.1 (70.3--75.8) & $+5.2$ (2.2--8.2) \\
Necrobinder$^{c}$ & raw & 1.4 & 1,000 & 63.1 (60.1--66.0) & 67.6 (64.6--70.4) & $+4.5$ (1.5--7.5) \\
\bottomrule
\end{tabular}

\smallskip
\parbox{0.92\linewidth}{\footnotesize
$^{\dagger}$These seeds were also used to choose the network or the search setting (Section~\ref{sec:eval-protocol}).\\
$^{a}$With the search after the fix of a leak (Appendix~\ref{sec:test-time-inference}); before the fix, the same seeds gave 68.4\% (66.7--70.0) with search, $+6.2$ points over greedy play.\\
$^{b}$On a separate set of 1,000 seeds, not used for any choice; on the 1,000 seeds on which the network was chosen, it won 68.4\% greedily.\\
$^{c}$On the first 1,000 seeds of the pool of Table~\ref{tab:retest}, drawn after all choices had been made. In its last 24 routine evaluations during training (updates 25,656--25,867, the same 192 seeds each time), the Necrobinder averaged 58.6\%.}
\end{table}

Table~\ref{tab:main-results} gives the main results.
With greedy play, the networks win 71--77\% of runs for the three characters of v0.111.0, and 62--68\% for the Fool, the Ironclad and the Necrobinder on v0.113.1.
Search in the two Act-3 boss fights adds 2.7 to 3.6 percentage points on the same seeds for the Defect, the Regent and the Silent (exact McNemar $p=3\times10^{-6}$, $0.002$ and $0.002$).
For the characters of v0.113.1, a different search method applied to all boss and elite fights adds 5.8, 5.2 and 4.5 points for the Fool, the Ironclad and the Necrobinder ($p=3\times10^{-11}$, $0.0008$ and $0.004$). The two gains are not directly comparable, but restricted to the same two Act-3 boss fights, the tree search gains about as much as the rollout search (below).

To check whether these choices inflated Table~\ref{tab:main-results}, we played the same networks with the same settings on the first 1,000 seeds of a pool drawn after all choices had been made (Table~\ref{tab:retest}).
On these seeds, the greedy win rates were within two points of Table~\ref{tab:main-results}: 74.9\% for the Regent, 71.2\% for the Silent, 62.5\% for the Fool (60.4\%, 58.6--62.1\%, on the first 3,000) and 66.5\% for the Ironclad.
The Defect, whose seeds had not been used for choices, won 77.8\% greedily on them, against 76.7\% in Table~\ref{tab:main-results}.
With the rollout search, the Silent's gain was 2.4 points ($p=0.001$), close to the 2.7 points of Table~\ref{tab:main-results}, and the Regent's was larger than there, 5.6 points against 3.1 ($p=2\times10^{-8}$).
The Fool's search gain was smaller, 4.2 points ($p=0.006$) against 5.8 with the same search on the original seeds. The two intervals overlap, but the difference is in the direction expected from choosing the search setting on the original seeds.

On the same seeds, we also ran the Fool's tree search in the two Act-3 boss fights only, the fights in which the rollout search runs.
It added 2.4 points (0.4--4.4, $p=0.03$), within the range of the rollout search's gains, and extending it to all boss and elite fights added another 1.8 points ($-1.1$ to $+4.7$, $p=0.25$) at about twice the time per game.

\begin{table}[t]
\centering
\small
\caption{Retests of the networks of Table~\ref{tab:main-results} on the first 1,000 seeds of a pool drawn after all choices of networks and settings had been made, with the same settings.
Win rates (\%) with 95\% Wilson intervals; $\Delta$ as in Table~\ref{tab:main-results}.
The last column repeats the greedy and search win rates of Table~\ref{tab:main-results}.}
\label{tab:retest}
\smallskip
\begin{tabular}{lcccc}
\toprule
Character & Greedy & Search & $\Delta$ & In Table~\ref{tab:main-results} \\
\midrule
\multicolumn{5}{l}{\emph{v0.111.0; rollout search in the two Act-3 boss fights}} \\
Defect   & 77.8 (75.1--80.3) & -- & -- & 76.7 / 80.3 \\
Regent   & 74.9 (72.1--77.5) & 80.5 (77.9--82.8) & $+5.6$ (3.7--7.6) & 74.1 / 77.2 \\
Silent   & 71.2 (68.3--73.9) & 73.6 (70.8--76.2) & $+2.4$ (1.0--3.8) & 71.3 / 74.0 \\
\midrule
\multicolumn{5}{l}{\emph{v0.113.1; information-set tree search in all boss and elite fights}} \\
Fool     & 62.5 (59.5--65.4) & 66.7 (63.7--69.6) & $+4.2$ (1.3--7.1) & 62.2 / 68.0 \\
Ironclad & 66.5 (63.5--69.4) & -- & -- & 67.9 / 73.1 \\
\bottomrule
\end{tabular}
\end{table}

In the Training column of Table~\ref{tab:main-results}, the Defect's 6.7 billion decisions include the development of the method; the other five characters were trained from scratch for 1.3 to 2.1 billion.

\begin{figure}[t]
\centering
\includegraphics[width=\linewidth]{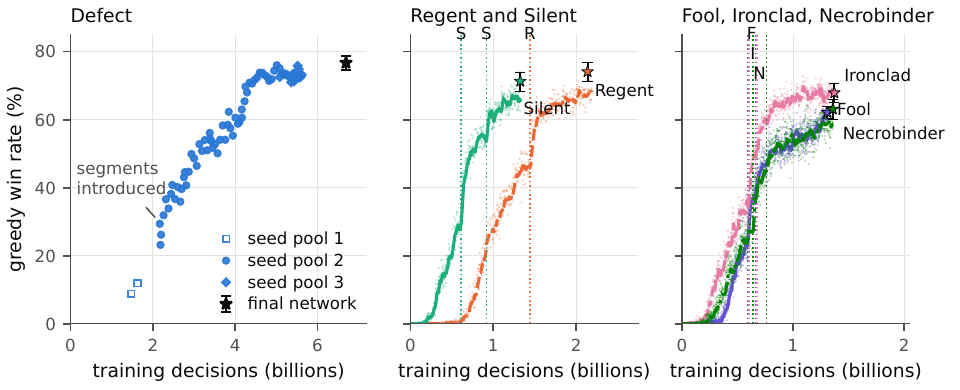}
\caption{Greedy win rate at Ascension~10 against the number of training decisions.
Left: the Defect (v0.111.0); each point is a test of one checkpoint on 1,000 fixed seeds, with marker shapes for the three seed pools used in turn.
Middle and right: the Regent and the Silent (v0.111.0), and the Fool (v0.113.0 until update 23,982), the Ironclad and the Necrobinder (v0.113.1); lines are trailing moving averages of routine evaluations of the raw weights (for the Fool from update 20,455, 0.83 billion decisions, of its schedule-free averaged weights) over 4,608 games (12 rounds of 384 seeds, or 24 rounds of 192 seeds for the Fool and the Necrobinder), and light dots are single rounds.
Dotted vertical lines, labeled with each character's initial, mark where the learning rate was halved and the batch doubled.
The Ironclad's curve ends at the network of Table~\ref{tab:main-results}.
Stars: the networks of Table~\ref{tab:main-results} on their test seeds, with 95\% Wilson intervals; the Defect's distilled network is placed at the decision count of its starting checkpoint.
The stars are EMA, distilled or (for the Necrobinder) raw networks on test seeds, while the curves show raw or, for part of the Fool's curve, averaged weights on routine-evaluation seeds, so the gap between them is not a measured gain.}
\label{fig:learning-curves}
\end{figure}

Figure~\ref{fig:learning-curves} shows how the greedy win rate grew with training.
In routine evaluation, the Silent passed 60\% after about 0.95 billion decisions and the Regent after about 1.5 billion; the Defect's 1,000-seed tests first exceeded 60\% at about 3.8 billion.
The Ironclad passed 60\% after about 0.75 billion decisions and the Fool after about 1.27 billion (on v0.113.0), while the Necrobinder ended its training just below 60\%.
The characters and the evaluation protocols differ, so these speeds do not isolate the effect of the recipe.
For the Regent and the Silent, the largest single gains followed a halving of the learning rate.

The Defect network was never trained on v0.113.0 or v0.113.1: after its first 0.3 billion decisions on v0.107.1, it was trained on v0.111.0.
On v0.113.1, after its vocabulary and content table were extended to the new version, it wins 76.9\% (74.2--79.4\%) of 1,000 greedy runs.
Its 76.7\% on v0.111.0 was measured on different seeds, so the difference, $+0.2$ points (95\% interval $-3.2$ to $+3.6$), is not paired.
On the first 1,000 seeds of the pool of Table~\ref{tab:retest}, it wins 74.8\% (72.0--77.4\%) greedily and 81.2\% (78.7--83.5\%) with the information-set tree search in all boss and elite fights, a gain of 6.4 points (3.6--9.2, $p=8\times10^{-6}$).
On one of these seeds, an engine used by the search failed reproducibly while copying the game state; we count this run as lost, although the network won it greedily.
Without this seed, the search wins 81.3\% of 999 runs and adds 6.5 points.
In its simulations, rewards follow the convention under which this network's value head was trained, with the progress rewards paid at the win (Appendix~\ref{sec:app-recipe}).

\subsection{Ablations of the Observation Encoding}
\label{sec:ablation}

To test the two parts of the observation encoding (Section~\ref{sec:observation}), we continued the Ironclad's training from its checkpoint at 0.60 billion decisions (update 18,354, 33.6\% in routine evaluation) with the same recipe and code in four branches, one run each.
The control continued unchanged.
In the branch without entity identities, the vocabulary IDs of cards, relics, potions, powers, enemies and their moves, orbs, enchantments, encounters and events were replaced by the unknown ID at the network's input, so that the ID embeddings and the content-table rows no longer told these entities apart; the record structure (kinds, zones, owners), map node types, option categories, the character and all numerical features were kept.
In the branch without engine-computed consequences, the consequences of public information that the engine computes (Section~\ref{sec:observation}) were set to zero: the damage and block of cards after modifiers, the end-of-turn previews, the damage and kill previews of candidates, and the block, HP loss, remaining HP, energy and cards drawn that follow a candidate; enemy intents were kept.
For comparison with a component of the training recipe, a fourth branch played full runs only, without combat practice.
Two further branches, one with a single credit timescale and one without privileged value inputs, were stopped before the end of the schedule and are not reported.
All branches followed the original schedule, which halves the learning rate at 0.60 and 0.68 billion decisions, and we evaluated each at 0.68 and 0.74 billion decisions greedily on 1,000 fresh seeds that were used for nothing else, paired seed by seed with the control.

\begin{table}[t]
\centering
\small
\caption{Ablations on the Ironclad (v0.113.1), each branch continued from the same checkpoint at 0.60 billion decisions. Greedy win rates (\%) on 1,000 fresh seeds at 0.68 and 0.74 billion decisions; difference from the control on the same seeds at 0.74 billion decisions, with a paired 95\% interval; runs reaching Act~3 (\%) at 0.74 billion decisions. One run per branch.}
\label{tab:ablation}
\smallskip
\begin{tabular}{@{}>{\raggedright\arraybackslash}p{30mm}ccccc@{}}
\toprule
 & \multicolumn{4}{c}{Raw weights} & \multicolumn{1}{c}{EMA weights} \\
\cmidrule(lr){2-5}\cmidrule(l){6-6}
Branch & 0.68 & 0.74 & Difference & Act~3 & 0.68 / 0.74 \\
\midrule
Control & 47.9 & 57.3 & -- & 85.4 & 55.7 / 57.6 \\
No entity identities & 0.0 & 0.0 & $-57.3$ ($-60.4$ to $-54.2$) & 31.0 & 0.0 / 0.0 \\
No engine-computed consequences & 43.0 & 49.0 & $-8.3$ ($-12.0$ to $-4.6$) & 82.1 & 20.4 / 38.2$^{a}$ \\
No combat practice & 50.8 & 58.1 & $+0.8$ ($-3.0$ to $+4.6$) & 86.3 & 54.8 / 60.5 \\
\bottomrule
\end{tabular}

\smallskip
\parbox{0.92\linewidth}{\footnotesize $^{a}$These EMA weights still average in weights trained before the fork with the full inputs, about a third of their weight at 0.68 and a fifth at 0.74 billion decisions, which lowers their win rate (see the text).}
\end{table}

\begin{figure}[t]
\centering
\includegraphics[width=\linewidth]{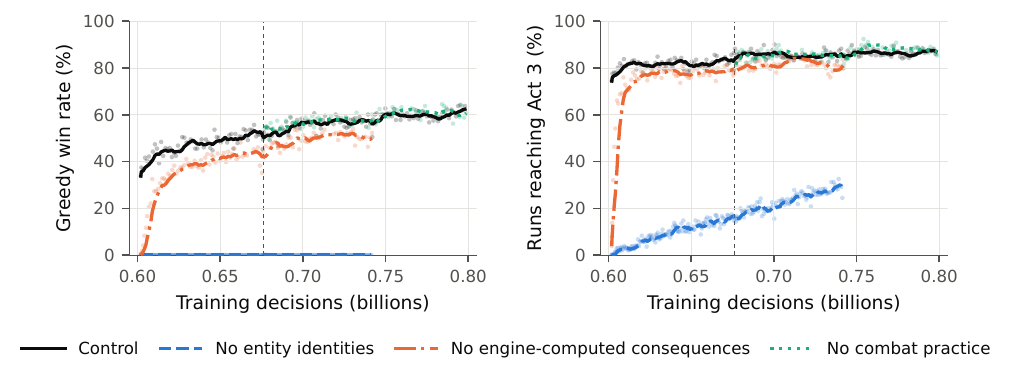}
\caption{Routine evaluation of the ablation branches (raw weights, 384 fixed seeds per evaluation): win rate and share of runs reaching Act~3 against training decisions. Lines are moving averages of six evaluations and dots single evaluations; the dashed line marks the second halving of the learning rate. The branch without combat practice has no routine evaluations before 0.68 billion decisions.}
\label{fig:ablation}
\end{figure}

Table~\ref{tab:ablation} and Figure~\ref{fig:ablation} give the results.
Without entity identities, the network did not win a single run: none of the 1,000 test runs at either reading, with raw or EMA weights, and none of the 86,400 runs of its routine evaluations after the fork.
It learned only to survive longer, reaching Act~3 in 31\% of test runs at 0.74 billion decisions against 85\% for the control.
After 0.14 billion decisions without identities, it reached Act~3 in 28\% of routine evaluation runs, less often than the original network had at the same number of decisions from scratch (57\%), when that network also won only 0.5\% of its runs; since the branch ran with the halved learning rates and larger batches of the late schedule and started from a network that relied on identities, this shows that the network cannot play without them, not how long a network would take to learn without them.
Without the engine-computed consequences, the network first lost almost every run and then recovered most of the gap: at 0.74 billion decisions it won 8.3 points fewer than the control (exact McNemar $p=2\times10^{-5}$) while reaching Act~3 about as often (82\% against 85\%), so most of the remaining loss came in Act~3.
Without combat practice, the network won as many runs as the control ($+0.8$ points at 0.74 billion decisions, $p=0.72$; $+2.9$ at 0.68, $p=0.16$), so in this late phase of training the practice episodes made no measurable difference.

We had planned to compare EMA weights, but the EMA of a branch whose inputs changed at the fork still averages in weights trained before the fork with the full inputs, which lowers its win rate: with EMA weights the branch without engine-computed consequences won 35.3 and 19.4 points fewer than the control at the two readings, a gap that shrank as the share of these weights fell.
We therefore also evaluated the raw weights of every branch, with the same protocol, after seeing the EMA results, and compare the branches on them; the routine evaluations of Figure~\ref{fig:ablation}, which use raw weights, agree with them.

These ablations remove a part of the encoding from a network that had been trained with it for 0.60 billion decisions and continue training for 0.14 billion more; they do not show how a network trained without that part from the start would fare, and each branch was run once, on one of two machines.

\subsection{Play in the Released Game}
\label{sec:live-results}

Five of the networks of Table~\ref{tab:main-results}, those of the Defect, the Regent, the Silent, the Fool and the Ironclad, and for the Silent also an earlier checkpoint, played the released game on a public livestream, through a mod that passes the game state to the policy and applies the chosen actions; no human made decisions in these runs.
These runs use the game's own random seeds and are not paired with the headless evaluations.
In the released game, each search was also limited to 4~s (rollout search) or 30~s (tree search), after which the greedy choice was played; the limit was rarely reached.
They test whether the results carry over from the headless engine to the released game.

\begin{table}[t]
\centering
\small
\caption{Runs at Ascension~10 in the released game, played through a mod.
Every completed run in each period is counted, including runs resumed after a crash, except as noted.
Expected: wins expected from the headless win rate of the same network and setting (Table~\ref{tab:main-results}; for the Silent's earlier checkpoint, 69.5\% with search on the Silent's test seeds; for the Fool's first five runs, 68.4\% with the search before the leak fix). Win rates (\%) with 95\% Wilson intervals.}
\label{tab:live-results}
\smallskip
\begin{tabular}{llllrrrc}
\toprule
Character & Network & Search & Period (2026) & Runs & Wins & Expected & Win rate \\
\midrule
\multicolumn{8}{l}{\emph{v0.111.0}} \\
Defect   & distilled & none (greedy)      & Nov 5--6   & 131 & 100 & 100.5 & 76.3 (68.4--82.8) \\
Defect   & distilled & Act-3 bosses       & Nov 6--8   & 173 & 138 & 139.0 & 79.8 (73.2--85.1) \\
Regent   & distilled & Act-3 bosses       & Nov 9--11  & 151 & 117 & 116.6 & 77.5 (70.2--83.4) \\
Silent   & EMA$^{a}$ & Act-3 bosses       & Nov 11--12 &  96 &  78 &  69.0 & 81.3 (72.3--87.8) \\
\multicolumn{8}{l}{\emph{v0.113.1}} \\
Fool     & EMA       & bosses and elites$^{b}$ & Nov 17--21 & 250 & 162 & 169.9 & 64.8 (58.7--70.5) \\
Ironclad & EMA       & bosses and elites$^{c}$ & Nov 21--22 &  65 &  50 &  47.5 & 76.9 (65.4--85.5) \\
\bottomrule
\end{tabular}

\smallskip
\parbox{0.92\linewidth}{\footnotesize
$^{a}$Two EMA checkpoints of the same training run: 37 of 45 runs with an earlier one and 41 of 51 with the network of Table~\ref{tab:main-results}.
One exhibition run on a preset seed, which the stream does not count, is excluded.\\
$^{b}$The first five runs used a search engine from before the leak fix (Appendix~\ref{sec:test-time-inference}), and one run, resumed after a restart, was played without search.\\
$^{c}$Runs completed by 08:40 CET on November~22; the livestream continued after that. In one run, the search's copy of the fight stopped matching the game at the run's 216th decision, and the rest of that run was played without search.}
\end{table}

Table~\ref{tab:live-results} lists these runs.
Of the 867 logged runs, the excluded exhibition run included, 31 were resumed after a crash, and 28 of these were won.
Each row has only 65 to 250 runs, so each can detect only large differences from the headless results.
Pooled over all 866 runs, the networks won 645, against 642.5 expected from the headless win rates of the same networks and settings, a difference of $+0.3$ percentage points (95\% interval $-2.7$ to $+3.3$).
The rows differ in both directions, from 8 wins fewer than expected for the Fool to 9 more for the Silent.
The Regent row starts after a packaging fault was corrected: the first packaged checkpoint lacked its format metadata, so its card and relic embeddings were zeroed at load time.
The 28 runs played with it (11 wins) are excluded.

\subsection{Where Runs Are Lost}
\label{sec:loss-analysis}

Table~\ref{tab:losses} shows where greedy runs end in a loss, on seeds drawn after all choices had been made.
\begin{table}[t]
\centering
\small
\caption{Where greedy runs are lost: the room of the fight in which each lost run ended, as a percentage of the losses, on the first 1,000 seeds of the pool of Table~\ref{tab:retest}, played greedily by the network of Table~\ref{tab:main-results} (the Defect's on v0.111.0).}
\label{tab:losses}
\smallskip
\begin{tabular}{lrrrrrr}
\toprule
Character & Losses & Act-3 bosses & \quad second & Act-1 boss & Act-1 elites & Act-2 boss \\
\midrule
Defect      & 222 & 68 & 51 &  5 &  7 &  8 \\
Regent      & 251 & 63 & 45 &  7 &  7 & 11 \\
Silent      & 288 & 52 & 38 & 13 &  9 &  7 \\
Fool        & 375 & 54 & 39 & 15 &  4 & 12 \\
Ironclad    & 335 & 67 & 55 &  8 &  5 &  9 \\
Necrobinder & 369 & 47 & 34 & 10 & 11 & 13 \\
\bottomrule
\end{tabular}
\end{table}

The two Act-3 bosses, the last two fights of the run, account for 47--68\% of the losses of every character, and the second of them alone for 34--55\%.
The Fool and the Silent lose more often than the others at the Act-1 boss (15\% and 13\% of their losses), and the Necrobinder at the Act-2 boss and in Act-1 elite fights (13\% and 11\%).
The positions analyzed below come from the greedy Defect evaluation of Table~\ref{tab:main-results}, in which 92.8\% of runs reached the first Act-3 boss, 96.6\% of those won it and 85.6\% of these won the second; of the 349 losses, 241 (69\%) occurred in these two fights, 48 at the first and 193 (55\% of all losses) at the second.

To test whether these losses come from the deck or from the play, we replayed each of the 241 Defect positions, taken just before the first Act-3 boss, with the random outcomes of both fights drawn anew.
CombatSolver~\cite{combatsolver}, the perfect-information solver of Section~\ref{sec:training}, to which the new outcomes were revealed and which had 120~s per fight, won both fights from 191 of the 241 positions (79.3\%, 73.7--83.9\%).
Under the same new outcomes, the Defect's network won both fights from 98 positions (40.7\%, 34.7--47.0\%) with greedy play and from 112 (46.5\%, 40.3--52.8\%) with the rollout search of Table~\ref{tab:main-results}.
The solver's result per position survives for 147 of the 241 positions; for the other 94, only its total was kept, because the per-position output of its first pass, on a rented machine that was later returned, was lost.
On these 147 positions, where the solver won 81.6\% (79.3\% on all 241), it won both fights 39.5 percentage points more often than greedy play (31.3--47.6, exact McNemar $p\approx10^{-16}$) and 34.7 points more often than search (27.0--42.4).
On all 241 positions the gaps are 38.6 and 32.8 points; filling in the 94 missing pairs in every way consistent with the totals gives exact McNemar $p$ of at most $7\times10^{-16}$ and $5\times10^{-13}$.
In these runs, choices offered at the start of a fight (such as a card chosen through a relic) were fixed to the first option, as they were for the solver; when the network made them itself, it won 42.3\% greedily and 51.5\% with search.
The comparison is rough in other ways: the solver was stopped after 120~s, and healing between the two fights was counted for the network but not for the solver (this could change the result of at most 13 positions).
These positions are thus not lost causes for the network itself, which wins two fifths of them under new outcomes.
The gap of 35 to 40 points to the solver is what knowing the outcomes in advance, together with the solver's own search, is worth on these positions; it does not tell how much of it better play under uncertainty could recover.

\subsection{Comparison with Human Players and Other AI Agents}
\label{sec:comparison}

Community statistics give a reference point for human play.
Spire Codex aggregates runs uploaded by players~\cite{spirecodex}.
On v0.111.0 at Ascension~10, uploaded single-player runs were won 14.8\% of the time with the Defect (15,948 runs), 17.1\% with the Regent (16,303), 19.9\% with the Silent (27,025) and 16.4\% with the Necrobinder (19,028), in data retrieved on October~26, November~6, November~9 and November~17, 2026, respectively.
On v0.113.1, in data retrieved on November~21, uploaded runs were won 13.4\% of the time with the Fool (3,056 runs), 10.9\% with the Ironclad (1,961) and 11.5\% with the Necrobinder (707).
These players chose to upload their runs and span all levels of skill, so the figures describe typical rather than expert play.
Few players win most of their runs (Table~\ref{tab:strong-players}): among players with at least 20 uploaded single-player runs at Ascension~10 on v0.111.0, 0.9--5.1\% win at least 80\% of them, depending on the character.

\begin{table}[t]
\centering
\small
\caption{Players on Spire Codex with at least 10 or at least 20 uploaded single-player runs on v0.111.0 at Ascension~10 (runs uploaded through November~22, 2026), and how many of them won at least 80\% of their runs~\cite{spirecodex}.
AlphaDefect: greedy and search win rates of Table~\ref{tab:main-results}.}
\label{tab:strong-players}
\smallskip
\begin{tabular}{lcccc}
\toprule
 & & \multicolumn{2}{c}{Players winning $\ge$80\%} & \\
\cmidrule(lr){3-4}
Character & Players & $\ge$10 runs & $\ge$20 runs & AlphaDefect \\
\midrule
Defect      & 438 & 7 of 207 (3.4\%) & 3 of 124 (2.4\%) & 76.7 / 80.3 \\
Regent      & 412 & 7 of 185 (3.8\%) & 4 of 112 (3.6\%) & 74.1 / 77.2 \\
Silent      & 511 & 9 of 260 (3.5\%) & 8 of 156 (5.1\%) & 71.3 / 74.0 \\
Ironclad    & 525 & 5 of 281 (1.8\%) & 3 of 159 (1.9\%) & 67.9 / 73.1$^{*}$ \\
Necrobinder & 430 & 4 of 205 (2.0\%) & 1 of 116 (0.9\%) & 63.1 / 67.6$^{*}$ \\
\bottomrule
\end{tabular}

\smallskip
\parbox{0.92\linewidth}{\footnotesize
$^{*}$The networks for these characters were evaluated on v0.113.1.}
\end{table}

Other AI agents trained for STS2 report lower win rates at Ascension~10, on earlier versions of the game.
On v0.107.1, the search-free policies of Chen and Xie win 8.1\%, 12.5\%, 2.4\%, 15.6\% and 1.6\% of 2,560 runs each with the Defect, the Regent, the Silent, the Ironclad and the Necrobinder~\cite{chen2026zerosearch}.
For the same five characters, Spire Mentor's October~2026 measurements report 22.0\%, 6.2\%, 9.0\%, 27.2\% and 5.6\% over 500 runs each, with 8,000 tree-search simulations per combat decision~\cite{spirementor}.
The four other characters were measured on October~11--13, and the Ironclad was re-evaluated on October~26.
According to Chen and Xie, these runs used an engine version that still granted rewards after the first Act-3 boss; run by Chen and Xie in their own host on common seeds, Spire Mentor won 8.8\% on average~\cite{chen2026zerosearch}.
The closest of these systems, developed concurrently, is that of Chen and Xie: it also trains PPO policies on the game's compiled assembly in a headless host and restarts training episodes from states the learner has visited, allocated by where full runs are lost~\cite{chen2026zerosearch}.
Its released policies were trained for about 1.0 to 1.8 billion collected decisions per character, close to ours after the Defect, comparable to our budget.
Figure~\ref{fig:headline} shows these figures beside ours.

\section{Discussion}
\label{sec:discussion}

We have introduced AlphaDefect, a state-of-the-art AI agent for Slay the Spire 2 that achieves high win rates across multiple characters and at the highest difficulty of Ascension~10. 
It far outperforms existing AI agents as well as most human players.

The following are some key observations and insights derived from our experiments with AlphaDefect.
\begin{itemize}
\item \textbf{The networks depend most on knowing what each entity is and what public information implies.}
Every card, relic, enemy and other piece of content enters the network through its entity record: a learned embedding of its vocabulary ID, a learned projection of a fixed row of attributes from the game data, and numerical features, among them the consequences of public information that the engine computes (Section~\ref{sec:observation}).
When the Ironclad's training was continued without these identities, the network did not win a single run, either on the 1,000 test seeds or in 86,400 routine evaluation runs; without the engine-computed consequences, it first lost almost every run and then recovered to 8.3 points below the control, the remaining gap lying mostly in Act~3; without combat practice, it won as many runs as the control (Section~\ref{sec:ablation}).
Two earlier observations, neither of them a controlled test, agree with this.
First, the Defect's network, trained on v0.107.1 and v0.111.0 only, won 76.9\% of 1,000 greedy runs on v0.113.1 once the vocabulary (2,290 IDs in use on v0.111.0, 2,557 on v0.113.1) and the content table (1,072 columns, 1,195) were extended, against 76.7\% on v0.111.0 on different seeds (Section~\ref{sec:main-results}).
This is what the content table is meant to allow, although we did not compare it with a network without the table.
Second, the first packaged checkpoint of the Regent, whose card and relic embeddings were zeroed at load time by a packaging fault, won 11 of 28 livestream runs (39.3\%; 95\% Wilson interval 23.6--57.6\%), against 117 of 151 (77.5\%; 70.2--83.4\%) for the corrected checkpoint of the same version; these runs are not paired and were not planned (Section~\ref{sec:live-results}).
The ablations show how heavily the network relies on this information, not that our way of encoding it is the best one.
\item \textbf{Checkpoints win different seeds.}
Plain PPO on the Defect stopped improving at 72--74\% after about 134,000 updates (4.6 billion decisions).
Checkpoints with the same average win rate win different seeds: of 1,000 seeds, 783 were won by some but not all of 15 checkpoints taken over about 8,000 updates, and two checkpoints fewer than 50 updates apart disagree on 14\% of seeds.
If wins were independent at 73\%, two checkpoints would disagree on 39\% of seeds; the observed 14\% shows that outcomes depend strongly, but not entirely, on the seed.
Runs diverge at near-ties between the two best candidates, typically in an early fight, and the long deterministic run amplifies the difference.
This is related to policy churn, the rapid change of greedy actions between nearby updates reported for value-based deep reinforcement learning~\cite{schaul2022churn}.
Halving the learning rate halved the KL divergence per update but reduced this disagreement by only 0.5--1.5 points.
\item \textbf{Averaging checkpoints removes part of this noise.}
An ensemble of checkpoints spread over training averages out these disagreements.
For the Defect, a student distilled from such an ensemble with the architecture of its starting checkpoint won 3.5 points more than that checkpoint (4.0 on the seeds not used to choose it), the only gain after the Defect's plateau that held; the evaluated student, which also dropped one attention component, won 6.4 points more than its own starting checkpoint on fresh seeds (Appendix~\ref{sec:app-recipe}).
For the Regent, the distilled student was no better than the EMA network it started from ($-0.9$ points, 95\% interval $-4.0$ to $+2.2$).
For the Silent, the EMA network already matched the ensembles.
\item \textbf{Search helps, but its gains are hard to keep.}
Search added 2.7 to 5.8 points for every character with a paired greedy comparison (2.4 to 5.6 points on the fresh seeds of Table~\ref{tab:retest}, and 6.4 for the Defect's network on v0.113.1), yet four attempts to distill its choices into the policy did not retain the gain, and more simulations did not help (Appendix~\ref{sec:test-time-inference}).
Greedy evaluation may also understate progress: a later Fool network beat an earlier one by 1.7 points greedily (95\% interval $-0.3$ to $+3.7$) and by 2.9 points (1.0--4.8) when both used the same search.
\item \textbf{Losses concentrate at the two Act-3 bosses.}
In greedy play on fresh seeds (Section~\ref{sec:loss-analysis}), 47--68\% of each character's losses occur in these two fights, and for the Defect a perfect-information solver won both fights from 79\% of the positions before them at which the network had lost, against 41\% for the network itself under the same newly drawn outcomes.
During the Defect's training, the hypothesis that deck building was the bottleneck was tested in five ways (counterfactual labels for card picks, a bias toward deck archetypes, ranking picks by the value head, doubling the non-combat samples and a reward for deck value), none of which found a gain; forcing an archetype cost 6 to 12 points.
\item \textbf{We chose fidelity over speed.}
We relied on the game's own rules for the headless results to carry over to the released game, and they did (Section~\ref{sec:live-results}).
Our own Rust reimplementation of the combat rules, built on the rules core of Spire Mentor~\cite{spirementor} and separate from sts2-sim, ran 30--50 times faster and agreed with the game on 98--99\% of combat steps with random decks.
It was used briefly to collect training data and then withdrawn, and on Act-3 boss positions from real runs it diverged almost from the first step; we did not use it for training or evaluation afterwards, so we have no controlled comparison between training in the headless engine, which runs the game's own rules, and training in our Rust reimplementation.
Training in a simulator of one's own requires it to match the game frame by frame, giving the same observation (Section~\ref{sec:observation}) at every decision of a run, which takes much more work than matching the game on combat steps alone.
After such a frame-by-frame check of sts2-sim against the game, a rerun of the Silent's training in sts2-sim reached 69.3\% (66.4--72.1) in the game with the protocol of Table~\ref{tab:main-results}, against 71.3\% for the network trained in the game (Appendix~\ref{sec:app-sim-training}).
\item \textbf{Networks transfer across game versions, not across characters.}
The Defect's network kept its win rate on v0.113.1 (Section~\ref{sec:main-results}).
Played unchanged as the Necrobinder, the Fool's network won none of 192 runs, although it averaged floor 17, about the level the Fool had reached after 120 updates of its own training.
A warm start therefore seemed to offer little, and the Necrobinder was trained from scratch; we did not test a warm start.

\end{itemize}

The following summarizes future directions and potential improvements for AlphaDefect.
\begin{itemize}
\item \textbf{Better value estimates for the Act-3 bosses.}
A search that saw the true future gained only about three points more than the public-information search, and repeated searches already agree on their choices, which suggests that the value estimates (accuracy of the value function) limit the search.
One possible cause is that the value head never saw zeroed privileged inputs in training, although a diagnostic found its estimates equally accurate without them, and a combat value head retrained on public information only was more accurate offline but did not strengthen the Fool's search ($-1.3$ points, 95\% interval $-3.4$ to $+0.9$).
Value targets for the Act-3 bosses, which account for about half or more of every character's losses, are the most direct place to improve.
\item \textbf{Bringing search into training.}
Since distilling the search's choices did not retain its gain (Appendix~\ref{sec:test-time-inference}), longer or controlled runs of such a loop, in the manner of expert iteration~\cite{anthony2017expert}, remain to be tried, as does collecting full runs with search so that non-combat play adapts to searched fights.
\item \textbf{Comparable evaluation.}
Comparisons with other agents on the same game version and seeds would make the results easier to interpret.
\item \textbf{Harder settings.}
The Dread mode added in v0.113.0 is much harder than Ascension~10: the better of two Defect networks trained in it, warm-started from a standard-mode checkpoint and chosen on the same 1,000 test seeds, wins 16.8\% of them.

    \item \textbf{Play diversity:} We found that the agent tends to narrow down its strategies as training proceeds. In particular, after some point the Silent model rarely takes the Shiv strategy. We think this is because once it learns to win with one strategy, the other strategies look suboptimal to the AI.
\end{itemize}

\paragraph{Limitations.}
Each character was trained once, so we cannot separate the recipe from the luck of a single training run, and the recipe changed from one character to the next (Table~\ref{tab:recipe}).
Most components were not tested in isolation (Table~\ref{tab:evidence}), and the ablations of Section~\ref{sec:ablation} remove parts of the observation encoding late in a single training run of one character rather than training without them from the start.
Several test pools were also used to choose networks or search settings; on fresh seeds, the greedy win rates stayed within two points and the search gains within 2.5 points of those reported (Table~\ref{tab:retest}).
Our intervals cover the sampling of seeds but not the variation between checkpoints or training runs.
For the Defect at its plateau, the standard deviation between checkpoints, 1.2 points, was about as large as the sampling error of a 1,000-seed evaluation, about 1.4 points, so the uncertainty about what the recipe achieves is larger than these intervals.
The search has known residual leaks of hidden information, which we expect to be small but have not measured directly.
Finally, our comparison with strong players uses runs on v0.111.0, so for the Ironclad and the Necrobinder it compares with players of an earlier version than the one our networks were evaluated on, and it does not cover the Fool; the other agents we compare with were also evaluated on an earlier version of the game.

\section{Related Work}

\paragraph{Agents for the original Slay the Spire.}
Most agents for the original game (StS1) control the game through Communication Mod, which exchanges the game state and commands with an external process~\cite{communicationmod,spirecomm}, or play in sts\_lightspeed, a C++ reimplementation of the game designed to reproduce its random number generation exactly, which also provides a tree search over combat~\cite{stslightspeed}.
Bottled AI, a published full-run agent, uses no learning: run-level choices follow hand-written priorities and rules, and combat is played by searching over simulated sequences of moves; it reports win rates of 20--52\% at Ascension~0, depending on the character~\cite{bottledai}.
Learning-based work has mostly addressed parts of the game.
Iyer et al.~\cite{kuo2020aiplaytesting} train a reinforcement learning agent on a reimplemented boss fight with a fixed deck to support balance testing, Zelnia~\cite{zelnia2023deckbuilding} compares card-picking algorithms that build on card synergies or simulate future fights while combat is left to an existing mod, and AutoClad~\cite{autoclad} imitates a scripted combat agent with a neural network.
Agents for full runs are still weak at the lowest difficulty.
Oram~\cite{oram_stsml} combines a double DQN for combat with a separate network for run-level decisions in a reimplementation of the game and reports a best training average of 24.8 floors over a consecutive window of 100 runs, with four wins; no fixed-policy evaluation was reported.
In its June~2026 report, AscensionAI~\cite{chan2026ascensionai} trains the Ironclad with PPO on the live game, after a warm start from a heuristic, and records a training average of 15.1 floors over the last 500 runs at Ascension~0; its published fixed-seed evaluations recorded no wins.
In sts\_lightspeed, a small network trained with REINFORCE for the run-level decisions, with combat played by the simulator's tree search, raises the Ironclad's win rate at Ascension~0 from 6\% to 14\%~\cite{stsrlagent2026}.

\paragraph{Learned and search-based agents for Slay the Spire 2.}
Several recent systems target Slay the Spire 2 (STS2), most of them released as open-source software rather than as papers.
Spire Mentor trains a run-level policy with PPO and plays combat with Monte Carlo tree search at several thousand simulations per decision, and reports win rates for all five characters at five ascension levels~\cite{spirementor}.
AlphaSpire releases a policy-and-search system whose networks for run-level and combat decisions guide a tree search~\cite{alphaspire}.
The sts2-rl-research project trains separate combat and choice agents without search~\cite{sts2rlresearch}.
Chen and Xie~\cite{chen2026zerosearch} train zero-search policies that select every action with a single forward pass of a 6.6M-parameter entity transformer.
They run the game's compiled assembly in a headless host that replays a run exactly from its seed and actions, and use this to restart training episodes from visited states, allocated in proportion to where complete runs are lost.
On 12,800 held-out games at Ascension~10 (A10) their policies win 8.05\% averaged over the five characters, and at Ascension~0 they match the 80.3\% reported by Spire Mentor.
AlphaDefect achieves significantly higher win rates at Ascension~10 (A10) than the other systems mentioned above.

\paragraph{Language-model agents and benchmarks.}
Slay the Spire has also become a testbed for large language model (LLM) agents.
Bateni and Whitehead~\cite{bateni2024language} study LLMs as game-playing agents in a reduced version of the original game, and MiniStS provides a simplified testbed modeled after the game in which cards can modify the game rules, aimed at game-playing AI and procedural content generation~\cite{bateni2024minists}.
Xiao and Yang~\cite{xiao2024testers} use LLM agents on Wordle and Slay the Spire to measure game difficulty, and find that agent performance correlates with the difficulty reported by human players even though it stays below the average human.
Orak includes Slay the Spire among twelve video games in a benchmark for training and evaluating LLM agents~\cite{park2026orak}.
Du et al.~\cite{du2025bottomup} evaluate a bottom-up skill-evolution agent that plays Slay the Spire from raw screen images and mouse actions without game-specific prompts, and their agent clears 13 floors at Ascension~0.
SpireBench evaluates 20 LLMs on the complete original game at Ascension~20 and finds that frontier models match the average human player but remain far behind experts~\cite{lowe2026spirebench}.
Epoch AI's SpireClimb evaluates frontier models on both games at their highest difficulties through a computer-use harness~\cite{epoch2026spireclimb,ou2026epochsts}.
In the October~2026 report, the best models win up to 98\% of runs in the original game at Ascension~0, but at most 20\% at Ascension~20 with a Heart kill and 35\% in STS2 at A10 on v0.107.1, whereas top human players are reported to win at least 80\% at the highest difficulty of both games~\cite{ou2026epochsts}.
For STS2, AgenticSTS studies bounded, typed memory for LLM agents, mainly at Ascension~0~\cite{cheng2026agenticsts}, and DemoEvolve uses human demonstrations to guide the evolution of an LLM agent's harness, raising the mean floor reached by Ironclad at Ascension~0 from 18.17 to 28.83 on held-out seeds~\cite{che2026demoevolve}.
Bateni et al.~\cite{bateni2025rulesynergy} evaluate how well LLMs analyze synergies between the rules introduced by cards, a skill that deck building in these games requires.
Beyond Slay the Spire, CivBench evaluates tool-mediated LLM agents over the long horizon of Civilization~VI~\cite{andrews2026civbench}.
These works use STS as a benchmark for general-purpose models, whereas the agents above are trained specifically for the game.

\paragraph{Reinforcement learning and search in other games.}
Superhuman agents in board games, poker, and real-time strategy games were trained with large-scale, distributed self-play~\cite{silver2018alphazero,brown2019pluribus,vinyals2019alphastar,berner2019dota}, often by combining a learned policy and value function with search and distilling the search back into the network~\cite{anthony2017expert,silver2017alphagozero}.
Distributed actor--learner architectures such as IMPALA scale reinforcement learning through parallel experience collection and off-policy correction~\cite{espeholt2018impala}.
STS2 is a single-player game, so no self-play opponent is available, and the strength of an agent is its win rate against the game itself.
Closest to our setting among card games, TavernAgent~\cite{ma2026tavernagent} trains an agent for Hearthstone Battlegrounds with PPO in a purpose-built fast simulator, using about 13 million games, and reaches the level of top players.
In Balatro, another single-player roguelike card game, Periasami et al.~\cite{periasami2026balatro} train separate PPO policies for playing hands and for the shop in a fast reimplementation of the game; their agent reaches on average the third of eight antes, and BalatroBench evaluates LLM agents on the same game~\cite{balatrobench}.
Choosing cards one at a time to build a deck has also been studied as drafting in collectible card games.
Vieira et al.~\cite{vieira2020drafting} learn drafting policies with deep reinforcement learning in Legends of Code and Magic, a card game designed for AI research whose competitions are summarized by Kowalski and Miernik~\cite{kowalski2023locm}, and Tales of Tribute provides a competition on a deck-building card game~\cite{kowalski2023tot}.

\paragraph{Analyses of Slay the Spire.}
Earlier work on the original game analyzes its stochasticity and decisions without training a full agent.
Trojanowski and Andersson~\cite{trojanowski2021lucky} and Andersson et al.~\cite{andersson2023lucky} analyze run data to separate the effects of randomness and player skill, and Nam and Rerkjirattikal~\cite{nam2023randomness} study the impact of randomness on outcomes.
Bazzaz and Cooper~\cite{bazzaz2025maps} measure the entropy of the paths players take through the procedurally generated maps in 20,000 runs and find that winning runs have higher normalized entropy, and Porenius and Hansson~\cite{porenius2021pathing} train a neural network that predicts successful paths through the map.

\section*{Author Contributions}
Q.C. conceived and led the project: set its goals, chose among the approaches that were tried, approved every change to the training runs, decided which experiments to run and when to stop them, and ran the livestream.
AlphaDefect (the headless host, the training system, both searches and the evaluation tools) was implemented, and the training runs and evaluations were operated, almost entirely by AI coding assistants (mainly Claude Code, and Codex for side tooling) working under Q.C.'s direction; the same assistants drafted Sections~\ref{sec:observation}--\ref{sec:results}, the discussion of these results and Appendices~\ref{sec:test-time-inference}, \ref{sec:app-recipe}, \ref{sec:app-choices} and~\ref{sec:code}, which Q.C. reviewed and is responsible for.
T.Y. leads the development of sts2-sim and is one of the main contributors to CombatSolver: the rollout search runs in sts2-sim, and CombatSolver chose the starting states of the solver-filtered combat practice and gives the ceiling in Section~\ref{sec:loss-analysis}.
T.Y. also directed the check of sts2-sim against the game and the rerun of the Silent's training in it (Appendix~\ref{sec:app-sim-training}); the lock-step test, the host that runs sts2-sim behind the headless engine's protocol and the training run were implemented and operated by an AI coding assistant (Claude Code) working under T.Y.'s direction, which also drafted that appendix, which T.Y. reviewed and is responsible for.
J.K. set up the manuscript and wrote the abstract, the motivation and challenges in the introduction, the problem formulation of Section~\ref{sec:learning-problem} and the formal description of credit assignment in Section~\ref{sec:training}, and the related work, and developed a variant of the training that uses community run data, on an earlier snapshot of the system (Appendix~\ref{sec:human-run-data}).

\section*{Acknowledgments}
We thank Spire Codex for publishing statistics of the runs that players upload~\cite{spirecodex}, the other contributors to sts2-sim~\cite{sts2sim} and CombatSolver~\cite{combatsolver}, and the developers of Spire Mentor~\cite{spirementor}, whose MIT-licensed Rust rules core we used in our experiments with a Rust reimplementation of the combat rules.
Slay the Spire~2 is a game by Mega Crit; this work is not affiliated with or endorsed by Mega Crit.
We thank Haobo Huang (The Chinese University of Hong Kong) for sharing with us the agent he developed independently and its results.

\bibliographystyle{alpha}
\bibliography{sample}

\appendix

\section{Materials and Methods}

Sections~\ref{sec:observation} and~\ref{sec:training} describe the observations and network, the environment, the training procedure and the rewards, and Section~\ref{sec:eval-protocol} the evaluation protocol; the hyperparameters of each character and the history of the recipe are given in Appendix~\ref{sec:app-recipe}.
This appendix describes training with states reconstructed from human runs and the test-time search in detail.

\subsection{Using Human Run Data}
\label{sec:human-run-data}

Combat practice from saved states (Section~\ref{sec:training}) only starts from states that the agent itself has reached.
To expose the agent to a wider range of decks and situations, we also start practice episodes from states reconstructed from runs played by humans.
We call this \emph{community practice}.
The agent plays on from the game states that human players reached using its own policy.

\paragraph{Data source.}
Spire Codex is a community site that collects run files uploaded by STS2 players and publishes statistics on them~\cite{spirecodex}.
A run file records the character, ascension level, game version, seed, and outcome of the run, the final deck, relics, and potions, and, for every floor, the rooms visited and the changes to the player's state (cards added, removed, upgraded, and enchanted, relics and potions gained and lost, HP, maximum HP, and gold).
We used the runs published up to November 13, 2026, which comprise 2,768,599 runs, of which 417,571 were played on v0.111.0.

\paragraph{Filtering.}
We kept only runs that match our setting: v0.111.0, the Silent, A10, a single player, the standard game mode, no modifiers, and not abandoned.
We removed runs flagged as cheated and runs that contain any card, relic, potion, act, encounter, monster, or event outside the official game content (i.e., modded runs).
We further removed runs whose records cannot be explained by the game rules, such as non-standard starting rewards, gold gains that no purchase, potion, or relic accounts for, and intervals in which HP, maximum HP, and gold are all recorded as zero.
These manual checks removed less than 2\% of the remaining runs.
After filtering, 25,711 runs remain, including runs that were lost.

\paragraph{State reconstruction.}
A Spire Codex run file contains the inventory only at the end of the run.
We therefore reconstruct the state after each floor by walking backward from the final inventory and undoing the recorded changes floor by floor: removed cards are added back, upgrades and enchantments are reverted, and relics and potions are restored.
When a change cannot be undone unambiguously, for example when two copies of a card carry different enchantments and the record does not tell which one was modified, we discard that floor and all earlier floors of the run and keep only the later ones.
Some information cannot be recovered from the records at all: the draw order, the state of the random number generators, and the counters of relics.
A reconstructed state therefore consists of the act, the floor, the deck, relics, potions, HP, maximum HP, gold, and the history of visited rooms and events, and the remaining parts are initialized fresh from the run's original seed.
The resulting states are not an exact replay of the human run, but the agent faces the same deck and resources as the human player did.
From the 25,711 runs we obtain 659,267 floor states, from which we extract 331,697 states just before a boss, elite, or normal fight.
We verified the reconstruction by loading reconstructed states in the actual game engine and checking that HP, maximum HP, gold, potion slots, deck, relics, and potions match the records.

We also prepared subsets restricted to won runs with a high HP at the end of the run, as a proxy for runs won comfortably.
Winning runs with a final HP of at least 33 (about half of the reviewed wins) give 2,722 runs and 109,952 floor states, and those with a final HP of at least 44 (about one third) give 1,857 runs and 74,038 floor states.

\paragraph{Training with community states.}
In community practice, half of the PPO data comes from community states.
An episode either restarts a boss, elite, or normal fight from a reconstructed pre-fight state, or continues the run from a reconstructed floor state until the run ends (a \emph{continuation}).
For continuations, the starting floor is first drawn uniformly from floors 1--48, and then a run that has a reconstructed state at that floor is drawn.
The data are split as follows: 25\% from ordinary combat practice from the agent's own states, 10.1\% each from community boss and elite fights, 13.5\% from community normal fights, 16.2\% from community continuations, and 25\% from regular runs from the first floor.
When drawing community fights, we favor fights whose deck contains cards that the agent has rarely played recently, so that combat practice covers cards the agent would otherwise not learn to use.

\paragraph{Results.}

We have trained a Silent model in v0.111.0. Initially it was trained with the setting of winning runs with a final HP of at least 33. Once the model was strong enough, we removed the human data and continued training solely with the agent's own experience.
We obtained a model with a win rate of 72.7\%, with a 95\% confidence interval of 69.6\% to 75.5\%. Although the model's strength does not significantly exceed that of the model without human data, it diversifies the strategies employed by the agent. The Silent model that was trained solely with its own experience rarely uses shiv-related cards, whereas this model uses both shiv- and poison-related cards; it picks Phantom Blades and Accuracy, which are typically picked when building a shiv-focused deck, with probability around 30\% of the occasions.

\subsection{Test-Time Search}
\label{sec:test-time-inference}

This appendix details the two test-time searches summarized in Section~\ref{sec:search}.
For the characters of v0.111.0 we used a rollout search over sampled worlds in the two Act-3 boss fights (called PIMC in our code and records).
For the characters evaluated on v0.113.1, we used a closed-loop tree search over information sets~\cite{cowling2012ismcts} in all boss and elite fights.
It samples worlds as perfect-information Monte Carlo (PIMC) does~\cite{long2010pimc}, but it does not solve each sampled world with perfect information: the network plays on from each candidate using only its public observations, as in Monte Carlo rollout search~\cite{tesauro1996online}.
Both use the trained network as their only learned component, for the prior over candidates and the value of the positions reached, and the rollout search also for the moves played inside simulations.

\paragraph{Information available to the search.}
A search must not use information that the player does not have.
Each search starts from the current position, rebuilt so that the history already revealed is reproduced: the tree search replays the run from its seed in the headless engine and reaches the real position exactly, while the rollout search follows the game in a separate simulator, described below.
In each simulated world, whatever is not yet revealed is then resampled: the engine reseeds the run's random number streams and reshuffles the draw pile, while the hand, HP and everything else on screen stay the same.
In the tree search and in the Defect's rollout search, cards known to be on top of the draw pile are reshuffled too, which discards a little public information; the simulator builds used for the Regent and the Silent keep them in place.
The value head's privileged inputs are set to zero.
The value head was trained with these inputs present, but in a diagnostic on 359 Act-3 positions of the Defect, its estimates agreed equally well with the outcomes of greedy rollouts with and without them (correlation 0.92 in both cases).

We found and closed one leak.
Until November~17, 2026, tree-search simulations that reached the reward screen after a fight saw the true rewards (gold, potion drops and elite relics), because these are drawn from streams that the reseed did not cover.
After the fix, the Fool's search evaluation repeated on the same 3,000 seeds gave 68.0\%, a paired difference of $-0.4$ points from the earlier engine (95\% interval $-1.7$ to $+0.9$), so the leak changed the result by less than two points.
Some random streams are still not reseeded.
In the headless engine, these are three run-level streams (UpFront, UnknownMapPoint and TreasureRoomRelics) and the streams owned by individual monsters, encounters and events; in the rollout search's simulator, only the random state of monsters summoned during a fight is derived from the run's seed.
None of the streams left in the headless engine decides anything in a fight that has begun: enemies choose their moves from a run-level stream that is reseeded, the random effects of cards draw from run-level combat streams that are reseeded too, a monster's own stream only picks its appearance, an encounter's stream only picks its monsters when the fight is created, and the three run-level streams serve the map, unknown rooms and treasure rooms.
For comparison, a diagnostic tree search that saw the true future outright won 2.8 points more than the public-information search (95\% interval 0.0--5.5) on 618 Act-3 positions of the Fool, the positions on which $\beta$ (below) was also tuned; this measures what full foresight would add, not the size of a leak.

\paragraph{Rollout search in the Act-3 boss fights (v0.111.0).}
The rollout search was built from the start on sts2-sim~\cite{sts2sim}, a C\# reimplementation of the game's rules that copies a combat state in about 0.1~ms; one of us (T.~Yu) leads its development.
We patched sts2-sim for fidelity to v0.111.0.
For the Defect, the Regent and the Silent, the search runs in sts2-sim, while the moves themselves are played in the game.
A mirror instance follows the game, importing the game's save, and the search falls back to greedy play where the mirror disagrees with the game.
In the Defect and Regent evaluations, the mirror agreed on 99.9\% and 99.8\% of decisions, and 1.4\% and 1.3\% of the decisions at which a search was due fell back.
At a searched decision, the search considers at most the four most probable candidates, each with probability at least 0.02.
Within each round, all candidates are played in the same sampled worlds, so that the comparison uses common random numbers.
After the candidate, the network plays greedily for 12 more decisions and the combat value head, with the mode input set to combat practice, scores the resulting position; a fight that ends earlier is scored by its outcome.
Two rounds of sequential halving~\cite{karnin2013halving} allocate the worlds: four per candidate in the first round, then, with three or four candidates, eight more each for the greedy candidate and the best other candidate (with two candidates, four more each), at most 32 rollouts.
Every candidate, including those dropped after the first round, is then compared with the greedy candidate on their common worlds, and the search switches to the one with the largest mean paired difference in value, provided that this difference exceeds two standard errors.
A search takes 0.7--0.9~s on average on a GPU and, in the Defect and Regent evaluations, changed the greedy choice at 10.2\% and 11.1\% of searched decisions.

\paragraph{Information-set tree search in boss and elite fights (v0.113.1).}
For these characters, the search runs in the headless engine itself, extended with in-process state cloning: a snapshot copies every mutable object of the game state and a restore writes the copies back, in about 1~ms and 0.1~ms on an idle machine (16~ms and 1.4~ms under full load).
On 100,941 combat observations from 2,221 branch points and 25,767 observations of whole runs, continuing from a restored snapshot gave the same observations as replaying the run up to that point.
Each of 32 simulations restores the root, resamples the hidden state as described above, plays a root candidate and descends the tree.
A node is identified by its parent, the action and the public state after it (decision type; the player's HP, block, energy and hand; the enemies' identities, HP and block; and the set of candidates), so deterministic plays reach the same node in every world while draws and enemy turns branch into chance nodes.
Actions are matched across worlds by a semantic key, such as the card, its upgrade and cost, and the target, rather than by their index.
A newly reached node is evaluated with one forward pass, without rollouts: its value is the reward collected along the path plus the network's whole-run value of the new position, in units of the training reward (undiscounted; a lost run adds nothing further).
The policy's probabilities become the node's prior.
Below the root, actions are selected as in Gumbel MuZero~\cite{danihelka2022gumbel}, with $Q$ normalized to $[0,1]$ within each node.
At the root, sequential halving spends the budget on the four most probable candidates, with common random numbers within each round.
The final choice maximizes $\log\pi(a)+Q(a)/\beta$ over the candidates with at least four visits, with $\beta=0.06$.
This is a regularized form of policy improvement~\cite{grill2020mcts}: the search leaves the greedy candidate $g$ for a candidate $a$ only if $Q(a)-Q(g)>\beta\log\bigl(\pi(g)/\pi(a)\bigr)$.
We tuned $\beta=0.03$ on Act-3 positions played with the mode input set to combat practice and doubled it for full runs, where values are about twice as large; the doubled value was not tuned further.
Under full load a search takes about 0.15~s, and with search in boss and elite fights a Fool run takes about 40~s instead of 8~s.

\paragraph{Variants and diagnostics.}
On Act-3 positions of the Fool, compared with 32 simulations, 128 or 256 simulations changed the win rate by $-0.3$ and $-0.2$ points (95\% intervals $-3.0$ to $+2.3$ and $-2.9$ to $+2.6$), and rollouts at the leaves by $+1.0$ ($-1.8$ to $+3.7$); other values of $\beta$ and more root candidates did not do better.
Searching every fight instead of only boss and elite fights changed the Fool's win rate by $-0.7$ points ($-2.4$ to $+1.1$).
Two searches of the same positions with independent worlds gave candidate value differences that correlated at 0.89 and chose the same final action 94\% of the time, so 32 simulations already give stable choices.
Together with the small gain from foresight, this suggests that the value estimates, rather than the search budget or the hidden information, limit the search.
Distilling the search's choices into the policy did not retain the gain: twice for the Defect, offline for the Fool, and during the Fool's training, where a KL term toward the search policy over 619 updates changed the greedy win rate by $-0.1$ points ($-3.3$ to $+3.1$).
Gains measured on positions just before the Act-3 bosses ($+6.4$ points on 2,472 positions with an earlier Fool network, 95\% interval 4.9--7.8) were about twice those in full runs, probably because such positions are played with the mode input set to combat practice, in which the network's greedy play is weaker than in full runs.
Table~\ref{tab:main-results} therefore reports full runs only.

\section{Training Recipe: Hyperparameters and History}
\label{sec:app-recipe}

This appendix complements Section~\ref{sec:training}.
Table~\ref{tab:hyperparameters} lists the training settings of the six evaluated networks, read from the configuration stored in each checkpoint; for the two distilled networks, it gives the settings of their starting checkpoints.
The rest of the appendix gives details that the main text leaves out and records how the recipe developed while the characters were trained, in the order the Defect, the Regent, the Silent, the Fool, and then the Ironclad and the Necrobinder in parallel.

\begin{table}[!tbp]
\centering
\footnotesize
\setlength{\tabcolsep}{2.6pt}
\caption{Training settings of the evaluated networks (for distilled networks, of their starting checkpoints). A single entry applies to all six characters. GAE cut: where the combat GAE is cut besides segment ends (fight: at the end of each fight; map: at the next map decision after it). Combat-practice shares are percentages of practice episodes; the remaining practice episodes replay normal and elite fights from saved states or use random decks.}
\label{tab:hyperparameters}
\smallskip
\begin{tabular}{@{}>{\raggedright\arraybackslash}p{55mm}cccccc@{}}
\toprule
 & Defect & Regent & Silent & Fool & Ironclad & Necrobinder \\
\midrule
\multicolumn{7}{@{}l}{\emph{Policy optimization}} \\
$\gamma$; GAE $\lambda$; non-combat $\lambda_{\mathrm{n}}$ & \multicolumn{6}{c}{1; 0.95; 0.95} \\
PPO clipping parameter; epochs per update & \multicolumn{6}{c}{0.2; 2} \\
Target KL (stop after an epoch above 1.5 times it) & 0.02 & \multicolumn{5}{c}{0.03} \\
Value-loss coefficient & \multicolumn{6}{c}{0.5} \\
Entropy coefficient, combat / non-combat & \multicolumn{6}{c}{0.01 / 0.05} \\
Categorical value loss (relative to value loss); fight-outcome loss & \multicolumn{6}{c}{0.1; 0.25} \\
Future-readout loss (horizon 12 decisions) & -- & -- & -- & 0.05 & 0.05 & 0.05 \\
GAE cut & -- & fight & fight & map & map & map \\
\midrule
\multicolumn{7}{@{}l}{\emph{Data and optimizer}} \\
Decisions per update & 131,072 & 65,536 & 131,072 & 131,072 & 131,072 & 131,072 \\
Minibatch $\times$ gradient accumulation & 2,048$\times$16 & 4,096$\times$2 & 4,096$\times$2 & 4,096$\times$1 & 4,096$\times$4 & 2,048$\times$4 \\
Segment length (decisions); staleness limit (updates) & \multicolumn{6}{c}{64; 6} \\
Pieces of 1,000 or more decisions exempt from the limit & no & \multicolumn{5}{c}{yes} \\
Adam learning rate ($10^{-5}$) & 5.5 & 20 & 10 & 15$^{a}$ & 3.75 & 7.5 \\
Muon learning rate ($10^{-5}$) & -- & -- & 15 & 7.5$^{a}$ & 7.5 & 15 \\
Gradient-norm clipping & 2.0 & \multicolumn{5}{c}{1.0} \\
EMA decay per update & \multicolumn{6}{c}{0.999} \\
\midrule
\multicolumn{7}{@{}l}{\emph{Combat practice}} \\
Engines in full runs & 3/4 & \multicolumn{5}{c}{1/2} \\
Boss fights, Acts 1 / 2 / 3 (\%) & 5/21/52 & 8/19/33 & 6/12/34 & 8/18/26 & 10/17/28 & 7/11.5/33.5 \\
Other fights from saved states, Acts 1 / 2 / 3 (\%) & 0/0/12 & 4/15/6 & 8/15/10 & 8/15/10 & 8/15/10 & 8/15/10 \\
Random decks (\%) & 11 & 15 & 15 & 15 & 12 & 15 \\
\midrule
\multicolumn{7}{@{}l}{\emph{Rewards}} \\
Per floor; Act-2 boss; first Act-3 boss; win & \multicolumn{6}{c}{0.02; 0.5; 1.5; 4} \\
Act-1 boss & 0.5 & 0.5 & 0.1 & 0.1 & 0.1 & 0.5 \\
Progress rewards paid only at a win & yes$^{b}$ & no & no & no & no & no \\
Practice episode: win / loss; HP change (per maximum HP) & \multicolumn{6}{c}{$+1$ / $-1$; 1} \\
Practice episode: won boss fight; lost boss fight (per enemy HP removed) & \multicolumn{6}{c}{$+0.5$; 0.5} \\
Discarded potion & \multicolumn{6}{c}{$-0.1$} \\
\midrule
\multicolumn{7}{@{}l}{\emph{Limits on stalled games}} \\
Repeated-position cut; fight lost after turns without progress & -- & \multicolumn{5}{c}{yes; 8 (20 in evaluation)} \\
Steps per practice fight & 600 & 5,000 & 5,000 & 5,000 & 1,000 & 1,000 \\
Steps per fight within a full run & -- & -- & -- & -- & 1,000 & 1,000 \\
Steps per full run & 6,000 & \multicolumn{5}{c}{20,000} \\
\bottomrule
\end{tabular}

\smallskip
\parbox{0.95\linewidth}{\footnotesize $^{a}$Schedule-free Adam, with schedule-free averaging also applied to Muon (warm-up 640 gradient steps, $\beta_1=0.9$, $\beta_2=0.999$).\\
$^{b}$From update 164,606 of the Defect's training, the rewards per floor and for the Act-1, Act-2 and first Act-3 bosses were scaled down linearly to zero over 300 updates, and the withheld amount was paid at the end of a won run; a won run thus kept its return, and only lost runs earned less. The scale reached zero at update 164,906, 112 updates before the Defect's starting checkpoint (stored settings \texttt{milestone\_from} 164,606 and \texttt{milestone\_ramp} 300).}
\end{table}

\subsection{Further Details of the Recipe}

\paragraph{Network.}
Screens that ask for several cards are split into single choices followed by a confirmation.
The layout of observations was revised as features were added, and we call each revision an observation format (Table~\ref{tab:recipe}).
The engine's previews leave out draws and other random effects, and damage to a random target is spread over the possible targets or, when deciding whether an enemy dies, ignored.
Features for a new character's mechanics, such as next-turn poison damage for the Silent, fill feature columns that other characters leave empty, so the architecture stays the same.
Features added in later observation formats reach existing networks through zero-initialized projections, which leave their outputs unchanged until trained.
The Transformer has a feed-forward width of 1,024 and no positional encoding.
The global vector is the mean of the encoded records plus embeddings of the decision type and of the mode (full run or combat practice).
The future-readout head predicts, for each combat candidate, five quantities over the next 12 decisions: the advantage, the change in HP and in enemy HP, and whether the fight is won or lost.
It has its own Adam optimizer (learning rate $10^{-4}$), and its correction to the candidate's score is trained by the policy loss alone.
The Ironclad's network runs its encoder twice from update 24,280: a first pass without gradients computes the policy's candidate probabilities, which zero-initialized layers add to the records of the second pass.
The networks have 19.4 to 19.6 million parameters, 4.7 million of them in the encoder and most of the rest in embedding tables.
Besides the value heads, a categorical value head trained with a cross-entropy loss on smoothed targets (128 bins on $[-2.5, 8]$)~\cite{farebrother2024stop} and the fight-outcome head add auxiliary losses (Table~\ref{tab:hyperparameters}); for the Defect, the value used for advantages and by the search came from the categorical head's expectation late in training.

\paragraph{Advantages and submission.}
With the cut at the next map decision, used from the Fool on, each batch also removes a systematic offset at the ends of fights in full runs.
For each act, the mean advantage of the last decision of a fight, computed up to the cut, is subtracted from the advantages and value targets of that fight's decisions, weighted by $(\gamma\lambda)^k$ for a decision $k$ steps before the last one.
The offset comes from a gap between the combat and non-combat value heads at the end of a fight, which otherwise favors ending fights earlier or later.
The cut itself had been at the first decision after a fight for the Regent and the Silent; it was moved to the map decision because non-combat values on the reward screens right after a fight were locally low, which made finishing a fight look worse than prolonging it.
Stale decisions that the learner keeps carry the action probabilities under which they were collected, and the PPO ratio is computed against them.
The learner computes in bfloat16; Adam uses $\epsilon=10^{-5}$ and no weight decay, and Muon~\cite{jordan2024muon} uses momentum 0.95 (Nesterov), five Newton--Schulz steps and updates scaled by $0.2\sqrt{\max(m,n)}$ for an $m\times n$ matrix.
In the Silent's network, Muon updates 44 hidden weight matrices, about 43\% of the parameters.

\paragraph{Sampling of practice episodes.}
Of the boss fights, 75\%, 75\% and 100\% in Acts 1, 2 and 3 start from saved states and the rest from generated decks of end-of-act size.
In Act~3, 30\% of the boss fights replay the second boss alone from the state saved before it.
Of the other boss fights from saved states, 30\% start from a state saved at a random map decision of the same act (each map decision's state is kept for this with probability 0.2) and fight one of the act's recently played bosses; the rest replay the boss fight that followed the saved state, in Act~3 as a pair of both bosses.
For the pairs, $\hat p$ is the win probability that corresponds to the rank of the value head's estimate at the start of the pair among the positions in the pool, through a calibration measured on 2,100 Act-3 positions of the Defect that were each replayed with 8 new seeds; positions ranked in the lowest 15\% are not drawn, and 10\% of the draws are uniform.
Second-boss positions are drawn with a weight proportional to the square of the recent loss rate in combat practice of their encounter.
Normal and elite fights from saved states are 50\%, 73\% and 40\% elite fights in Acts 1, 2 and 3, and fights with random decks are 40\% elite fights (all elite fights for the Defect).
Each actor keeps the 256 most recent saved states of each kind.
A won practice boss fight earns 0.5 more, and a lost one earns 0.5 times the fraction of the enemies' starting HP that the policy removed.

\subsection{How the Recipe Developed}

\paragraph{The Defect.}
The recipe was developed on the Defect, whose training differed most from the final form.
At first, runs were sent to the learner only after they ended, and only about a third of the decisions from the first act reached the learner within the staleness limit, which was then three updates.
The Defect's greedy win rate on the same 1,000 seeds rose from 12.6\% at a checkpoint about 1,200 updates before segments were introduced to 29.4\% at one about 1,100 updates after; this comparison also includes the effect of the training in between.
Its combat GAE was cut only at segment ends, it used three epochs per update until update 107,982, and attempts to carry the non-combat chain across segment ends were switched off again.
Its combat practice started with fights on generated decks whose difficulty followed the win rate in combat practice, which saved states then replaced, and the share of engines in full runs rose from 0.3 to 0.5 and then to three quarters.

Shaping terms used early in the Defect's training, such as those for HP, relics, upgrades and potion use, were removed in several steps, for two reasons: hand-designed shaping can change which policy is optimal, and a calibration against replayed boss fights found that the terms overvalued upgrades and relics by factors of two to six.
Late in the Defect's training the progress rewards were phased out over 300 updates and paid in full at a win instead.

Early in the Defect's training, the policy won only about 4\% of its practice episodes against the Act-2 boss.
For about two days, half of the practice episodes against bosses that were drawn from saved states therefore started from states that a perfect-information fight solver, CombatSolver~\cite{combatsolver}, had found winnable, with priority for states the policy had not yet won.
One of us (T.~Yu) is a main contributor to CombatSolver.
Half of these episodes replayed the random seed on which the solver had won.
The filter covered the Act-1 and Act-2 bosses and later the Act-3 bosses, and was switched off when the solver's machine was needed for other work.
This oracle affected only which states combat practice started from.
In its last 1,000 or so updates before the starting checkpoint of its student, the Defect also ran an extra phase every 32 updates that trained the value heads on stored data while keeping the policy close to its previous output; it was removed afterwards.

\paragraph{The Regent and the Silent.}
The Regent introduced the cut at the end of each fight, the rules that end stalled fights (Table~\ref{tab:hyperparameters}) and the exemption of long pieces from the staleness limit, which keeps fights that loop for thousands of decisions in training.
The Defect and the Regent used Adam alone; a Muon trial in the middle of the Regent's training, at a learning rate chosen to match Adam's policy change per update, lowered the full-run win rate in training by about 5 points and was rolled back.
From the Silent on, Muon was used from the first update.
On the Silent, the first two changes of the schedule rule were followed by rises in routine evaluation from 29\% to 47\% and from 53\% to 63\%, within about 1,300 and 500 updates.
At the end of the Silent's training, boss fights made up 52\% of practice episodes, about two thirds of them in Act~3.

Late in the Silent's training, three quarters of its full runs skipped easy fights: a normal or elite fight that the auxiliary head predicted to be won with probability at least 0.995 was skipped with probability 0.8, and the HP and potions after it were sampled from learned heads that predict them.
This raised the number of full runs per hour by a third without a measurable change in strength; no other character used it.
Skipping every such fight instead lowered the Silent's win rate in practice episodes against the Act-3 bosses and was withdrawn.

\paragraph{The Fool, the Ironclad and the Necrobinder.}
The Fool introduced the cut at the next map decision and, from update 22,975, the future-readout head; the Ironclad and the Necrobinder had both from the start.
From update 20,455, both of the Fool's optimizers used schedule-free averaging~\cite{defazio2024schedulefree}; a trial of it on the Ironclad was discarded.
The schedule rule halved the learning rates and doubled the batch twice for each of the three (at updates 18,104 and 19,161 for the Fool, 18,355 and 19,479 for the Ironclad, 19,450 and 21,310 for the Necrobinder) before further steps accumulated gradients or halved the learning rates alone.
For the Necrobinder, cutting the Act-1 boss reward to 0.1 was reverted together with the weights.
From update 23,699 to 24,056, the Ironclad used a control variate for the luck of chance events in boss and elite fights, which cost about 42\% of the collection throughput, and switched it off when an evaluation guardrail was reached; a later trial with extra HP in training was discarded.

\subsection{Ensembles and Distillation}

For the Defect and the Regent, the evaluated network is a student distilled~\cite{hinton2015distilling,rusu2016distillation} from an ensemble of checkpoints of the same training run.
The teacher averages the action probabilities of 16 (Defect) or 8 (Regent) checkpoints spread over the later part of training.
The student starts from a recent checkpoint and is trained for a few thousand gradient steps on about 262,000 positions recorded during training.
It minimizes $D_{\mathrm{KL}}(\pi_{\mathrm{teacher}}\,\|\,\pi_{\mathrm{student}})$ on decisions with more than one candidate, plus a squared-error penalty (weight 0.5) that keeps its value estimates close to those of its starting network.
For the Defect, another student (not the evaluated one) distilled from the same teacher, with the architecture of its starting checkpoint (update 156,880), won 3.5 percentage points more than that checkpoint on 3,500 test seeds (exact McNemar $p<10^{-4}$), about as much as the ensemble itself; 1,500 of these seeds were used to choose it over a second student, and on the other 2,000 the gain was 4.0 points ($p=0.0003$).
For the Regent, the teacher, an ensemble of 8 checkpoints, won 5.2 points more than the single checkpoint at update 53,840 on the 1,000 seeds that were also used to choose it ($p=0.003$).
During the project, neither evaluated student was compared with its starting checkpoint in full runs; we made these comparisons afterwards, greedily, on the first 1,000 seeds of the pool of Table~\ref{tab:retest} and on the seeds of Table~\ref{tab:main-results}, with the code and settings of Table~\ref{tab:main-results}.
The Defect's student won 77.8\% (75.1--80.3\%) against 71.4\% for its starting checkpoint, update 165,018 ($+6.4$ points, 95\% interval 3.1--9.7, exact McNemar $p=2\times10^{-4}$), and 76.7\% against 71.3\% on the 1,500 seeds of Table~\ref{tab:main-results} ($+5.4$, 2.7--8.1, $p=1\times10^{-4}$).
Besides distillation, this comparison includes one change of architecture: in non-combat decisions the starting network also attended over records of the deck, the relics and other run-level information, and the student did not.
The Regent's student won 74.9\% (72.1--77.5\%) against 75.8\% (73.0--78.4\%) for its starting point, the EMA weights of update 54,603 ($-0.9$, $-4.0$ to $+2.2$, $p=0.61$), and 74.1\% against 72.4\% on the 1,000 seeds of Table~\ref{tab:main-results}, on which its teacher had been chosen ($+1.7$, $-1.5$ to $+4.9$, $p=0.33$).
Its distillation thus kept, but did not add to, the strength of the network it started from.
On the Silent, ensembles did not significantly outperform the EMA network (at best by 0.8 points, $p=0.67$), and no student was distilled.

\section{Card and Relic Choices}
\label{sec:app-choices}

This appendix describes the choices that the networks make outside combat.
The data are the last 4,608 routine-evaluation games of each training run of Figure~\ref{fig:learning-curves}, played greedily at Ascension~10 with the raw weights at the end of training (for the Fool, with its schedule-free averaged weights).
For comparison, we use the card statistics that Spire Codex~\cite{spirecodex} publishes for single-player runs at Ascension~10 on the same game version, retrieved on November~21, 2026; they cover 707 runs for the Necrobinder, 1,961 for the Ironclad, 3,056 for the Fool and 20,361 to 30,861 for each character of v0.111.0.
The two sources differ in kind.
A network plays greedily, so in a given position the same offer always gets the same choice, and its pick rates average over positions only; the players' rates also average over players of all levels of skill.

\begin{figure}[ht]
\centering
\includegraphics[width=\linewidth]{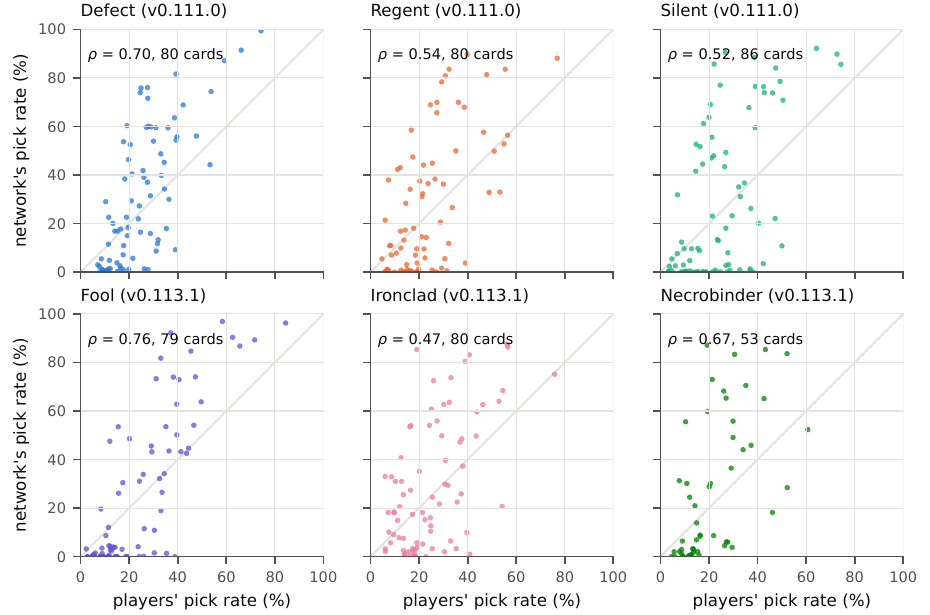}
\caption{Pick rates of individual cards on card-reward screens: the network, in the last 4,608 routine-evaluation games of its training run, against players on Spire Codex on the same game version.
Each point is a card offered at least 50 times to the network and 200 times to players; $\rho$ is the rank correlation, and the diagonal marks equal rates.}
\label{fig:card-picks}
\end{figure}

\paragraph{Card rewards.}
The networks saw 15 to 19 card-reward screens per run and took no card on 9 to 22\% of them: 4 to 12\% in Act~1 and 12 to 30\% in Acts~2 and~3 (Table~\ref{tab:choices}).
Over the cards offered often enough to both, the networks rank cards much as the players do (rank correlation 0.47 to 0.76), but their pick rates are more extreme: most cards are taken almost always or almost never (Figure~\ref{fig:card-picks}).
Table~\ref{tab:choices-cards} lists the largest differences.
The Silent's network, for example, almost never takes Knife Trap or Fan of Knives, two cards built around Shivs that players take in more than a third of offers, in line with the observation on play diversity in Section~\ref{sec:discussion}.

\paragraph{Relics.}
Relics from fight rewards and treasure rooms are almost always taken: the networks skipped 0 to 31 of the 2,159 to 4,379 such relics they were offered, at most 1.2\%.
At the start of each act, an Ancient offers a choice of relics.
Of the roughly 100 relics offered at least 100 times to each network, it chose 5 to 8 whenever they were offered and never chose 25 to 32 (Table~\ref{tab:choices-ancients}).
Some preferences are shared: three networks always took Big Mushroom, and Hefty Tablet, Wuppy and Fragrant Mushroom are each among the relics that at least three networks never took.
Spire Codex has no comparable records of players' choices among these relics.

\paragraph{Shops and rest sites.}
Per run, the networks bought 2.6 to 3.8 cards and removed 1.1 to 1.5 cards at shops, and at rest sites they upgraded a card at 44 to 63\% of the visits at which they upgraded or healed.
They took 7.9 to 9.2 potions per run and discarded 1.4 to 2.3, which our engine allows only when a slot is needed for a new potion.

\begin{table}[ht]
\centering
\footnotesize
\setlength{\tabcolsep}{4pt}
\caption{Choices outside combat in the last 4,608 routine-evaluation games of each training run. Screens: card-reward screens per run. Skipped: share of those screens on which no card was taken, in Acts~1, 2 and~3 (\%). $\rho$: rank correlation between the network's and the players' pick rates over the cards offered at least 50 times to the network and 200 times to players (number of cards). Shop: cards bought and cards removed per run. Upgrade: upgrades as a share of upgrades and heals at rest sites (\%). Potions: taken and discarded per run.}
\label{tab:choices}
\smallskip
\begin{tabular}{lccccccc}
\toprule
 & Screens & Skipped & $\rho$ & Shop buys & Removals & Upgrade & Potions \\
\midrule
Defect & 19.1 & 4, 12, 14 & 0.70 (80) & 3.8 & 1.1 & 46 & 8.3, 1.5 \\
Regent & 18.3 & 12, 26, 28 & 0.54 (80) & 2.8 & 1.1 & 45 & 8.4, 2.3 \\
Silent & 15.2 & 12, 19, 18 & 0.52 (86) & 3.2 & 1.5 & 60 & 7.9, 1.5 \\
Fool & 16.4 & 12, 30, 29 & 0.76 (79) & 3.4 & 1.5 & 44 & 8.6, 1.5 \\
Ironclad & 16.4 & 10, 30, 30 & 0.47 (80) & 2.6 & 1.3 & 63 & 9.2, 1.5 \\
Necrobinder & 16.2 & 11, 25, 26 & 0.67 (53) & 3.4 & 1.4 & 62 & 7.9, 1.4 \\
\bottomrule
\end{tabular}
\end{table}

\begin{table}[ht]
\centering
\footnotesize
\caption{Cards that the networks pick much more or much less often than players, among the cards compared in Table~\ref{tab:choices}: the network's pick rate / the players' (\%).}
\label{tab:choices-cards}
\smallskip
\begin{tabular}{>{\raggedright\arraybackslash}p{0.14\linewidth}>{\raggedright\arraybackslash}p{0.38\linewidth}>{\raggedright\arraybackslash}p{0.38\linewidth}}
\toprule
 & Picked more often & Picked less often \\
\midrule
Defect & FTL 76/25; Go for the Eyes 74/25; Thunder 76/28 & Trash to Treasure 9/39; Darkness 1/28; Genetic Algorithm 1/26 \\
Regent & Spectrum Shift 84/32; Make It So 81/31; Manifest Authority 90/40 & Decisions, Decisions 4/39; Sword Sage 0/31; Royal Gamble 2/29 \\
Silent & Coolheaded 91/27; Leap 86/22; Bubble Bubble 77/25 & Blade of Ink 11/50; Expose 4/42; Knife Trap 0/38 \\
Fool & Motif 92/37; Ki Rest 82/33; Ki Whip 73/31 & Quiver 0/39; Yah! 1/35; Rapid Shot 2/30 \\
Ironclad & Stone Armor 85/19; Aggression 72/26; Uppercut 83/41 & Dark Embrace 1/41; Bloodletting 21/54; Second Wind 3/35 \\
Necrobinder & Shroud 87/19; Sleight of Flesh 83/30; Countdown 73/21 & Lethality 18/46; Capture Spirit 4/30; Delay 28/52 \\
\bottomrule
\end{tabular}
\end{table}

\begin{table}[ht]
\centering
\footnotesize
\caption{Relics chosen from the Ancients at the start of each act, over the relics offered at least 100 times to the network (number of relics): how many it always chose, how many it never chose, and examples of each.}
\label{tab:choices-ancients}
\smallskip
\begin{tabular}{lccc>{\raggedright\arraybackslash}p{0.27\linewidth}>{\raggedright\arraybackslash}p{0.27\linewidth}}
\toprule
 & Relics & Always & Never & Always chosen, e.g. & Never chosen, e.g. \\
\midrule
Defect & 103 & 6 & 28 & Toasty Mittens, Pael's Blood, Alchemical Coffer & Kaleidoscope, Hefty Tablet, Pael's Eye \\
Regent & 103 & 7 & 26 & Big Mushroom, Small Capsule, Lava Rock & Fragrant Mushroom, Touch of Orobas, Tri-Boomerang \\
Silent & 103 & 5 & 30 & Sai, Seal of Gold, Nutritious Oyster & Hefty Tablet, Leafy Poultice, Dowsing Rod \\
Fool & 101 & 8 & 32 & Archaic Tooth, Music Box, Pael's Blood & Wuppy, Hefty Tablet, Scroll Boxes \\
Ironclad & 103 & 5 & 25 & Big Mushroom, Large Capsule, Toasty Mittens & Fragrant Mushroom, Wuppy, Winged Boots \\
Necrobinder & 102 & 8 & 28 & Big Mushroom, Touch of Orobas, Fiddle & Wuppy, Hefty Tablet, Fragrant Mushroom \\
\bottomrule
\end{tabular}
\end{table}

\clearpage

\section{Code}
\label{sec:code}

The code of the system described in this paper, all trained networks and the records behind our results are available under the MIT license at \url{https://github.com/hmty-CHEN/AlphaDefect-Public}.

\paragraph{Contents.}
The release contains:
\begin{itemize}
    \item the headless host, which runs the game's own assembly in a .NET~9 process, with its observation encoder and the vocabulary and content table that the networks use as input; there are versions for v0.111.0 and for v0.113.1, the latter with in-place state cloning (Appendix~\ref{sec:test-time-inference});
    \item the training system, written in Python with PyTorch: the learner, the rollout workers, practice episodes, weight averaging and ensemble distillation;
    \item both test-time searches, the rollout search, called PIMC in the code, with our patches to sts2-sim~\cite{sts2sim} and the information-set tree search;
    \item the evaluation scripts and the seed lists of Table~\ref{tab:code-networks}, including the list used for the retests of Table~\ref{tab:retest};
    \item the networks of Table~\ref{tab:main-results} and other checkpoints of the training runs, among them the Defect's distillation teacher pool, each with a manifest that records its observation format, training update, number of training decisions and checksum;
    \item the per-seed results of the evaluations in Tables~\ref{tab:main-results} and~\ref{tab:retest}, of the Defect's network on v0.113.1 (Section~\ref{sec:main-results}), of the distilled students against their starting checkpoints (Appendix~\ref{sec:app-recipe}) and the place of each loss in Table~\ref{tab:losses}, the run log of the released game behind Table~\ref{tab:live-results}, the training logs and routine evaluations behind Figure~\ref{fig:learning-curves}, the scripts that regenerate the figures and the tables of Appendix~\ref{sec:app-choices}, and the project's decision logs.
\end{itemize}

\paragraph{Reproducing the evaluations.}
Table~\ref{tab:code-networks} lists, for each row of Table~\ref{tab:main-results}, the released network and the seeds behind each row.
Running the host requires a copy of Slay the Spire~2 at the version of that row, the .NET~9 SDK and Python with PyTorch; greedy evaluation runs on a CPU.
Because beta versions of the game may not stay available, we identify each build by the SHA-256 of its assembly \texttt{sts2.dll}: \texttt{0861bfa1df347538\ldots} for v0.111.0 and \texttt{62ecbe91093e1fa8\ldots} for v0.113.1, with the full hashes in the release.
On the same hardware and batch size, greedy evaluations can be repeated game by game, because the host is deterministic given the seed and the actions, and greedy play is deterministic given the network.
In our checks, playing the same checkpoint twice on 32 seeds and twice on 384 seeds gave identical runs, a repeated evaluation of an earlier Silent checkpoint reproduced the outcome and the floor reached on all 1,000 of its test seeds, and two machines with different GPUs gave identical results on each of 400 seeds.
Rerun on the training machine one to two weeks later, the greedy evaluations behind the Silent's and the Regent's rows of Table~\ref{tab:main-results} gave the same outcome in every game: all 1,000 of the Silent's, both with and without the rules that end stalled games, and all 1,000 of the Regent's.
On another machine, with RTX~5090 GPUs and the engine rebuilt from the same source, the greedy evaluations behind the Defect's and the Regent's rows of Table~\ref{tab:main-results} again gave the same outcome in every game (1,500 and 1,000), as did the Regent's retest of Table~\ref{tab:retest}.
Across devices, rare near-ties between candidates can flip a decision: replayed on a CPU, 45 of 48 runs of the Ironclad's two networks from their original GPU test ended the same way (outcome, floor reached and number of fights).
In observation formats up to 28, one feature of one event, the entrant number of the Trial, is also drawn from a random stream that the seed does not fix.
The two rules that end stalled games are options of the evaluation scripts (Section~\ref{sec:eval-protocol}).
Each character after the Defect was trained from scratch in two to four days, with one learner GPU and a few hundred engine processes (Section~\ref{sec:training} and Table~\ref{tab:recipe}).

\begin{table}[ht]
\centering
\small
\caption{Released networks and test seeds for the rows of Table~\ref{tab:main-results}. Networks are identified by the training update at which they were saved. W1, P2-future, R1 and RV1 are seed lists in the release; entries are counted from 1. The retests of Table~\ref{tab:retest} also use the first 1,000 entries of RV1 (the first 3,000 for the Fool's greedy retest).}
\label{tab:code-networks}
\smallskip
\begin{tabular}{llll}
\toprule
Character & Game & Network & Test seeds \\
\midrule
Defect & v0.111.0 & distilled (student~e) & W1, entries 6,901--8,400 \\
Regent & v0.111.0 & distilled (student~e8) & W1, first 1,000 \\
Silent & v0.111.0 & EMA, update 26,428 & W1, first 1,000 \\
Fool & v0.113.1 & EMA, update 24,181 & P2-future, first 3,000 \\
Ironclad & v0.113.1 & EMA, update 24,782 & R1, all 1,000 \\
Necrobinder & v0.113.1 & raw, update 25,875 & RV1, first 1,000 \\
\bottomrule
\end{tabular}
\end{table}

\paragraph{Evaluation records.}
Complete records of several greedy evaluations, with every decision, the candidates offered and the probability that the network gave each of them, are already public at \url{https://github.com/hmty-CHEN/alphadefect-evaluation-runs}.
They cover 1,000 games of the Silent's network of Table~\ref{tab:main-results} on its test seeds, and 1,000 games each of other checkpoints of the Defect, the Regent and the Fool.
The Silent's games there were played with the rules that end stalled games, and 712 of them were won, against 713 in Table~\ref{tab:main-results}.

\paragraph{License and third-party material.}
The MIT license covers our own work only.
Slay the Spire~2 is a game by Mega Crit, and this work is not affiliated with or endorsed by Mega Crit.
The release contains no game binaries, art, audio or decompiled source code.
It does contain the vocabulary and content table, which list entity IDs and numerical attributes derived from the game's data and which the networks need as input; these are not covered by the license.
Our patches to sts2-sim are released as patch files; sts2-sim itself, which has its own MIT license, and CombatSolver~\cite{combatsolver} are separate projects and are not included.
Neither is the mod that played the released game in Section~\ref{sec:live-results}.

n
\section{Training in a Reimplementation of the Game}
\label{sec:app-sim-training}

All networks in the main text were trained in the headless engine, which runs the game's own rules (Section~\ref{sec:training}).
One of us (T.~Yu) repeated the Silent's training from scratch with sts2-sim~\cite{sts2sim}, a reimplementation of the rules, in place of the game, and evaluated the resulting checkpoints both in sts2-sim and in the game.
The purpose was to test whether a reimplementation can replace the game as the training environment once it has been checked against the game frame by frame.
The experiment is a single training run, and it was made after the results of the main text; none of them depends on it.

\paragraph{Checking the simulator against the game.}
A second host runs sts2-sim (its v0.111.0 line) behind the protocol of the headless engine, with a port of the same observation encoder.
A lock-step test then plays one run in both hosts at once: the same seed, and at every decision the same action, chosen greedily by the Silent's network of Table~\ref{tab:main-results} from the game's frame.
At every decision the two frames, that is, the observation and the list of candidates, are compared, and the run stops at the first difference.
On the first 1,000 seeds of W1, 647 runs matched the game at every decision from the first floor to the end when the test was first run; 346 stopped at a difference in the state and 7 at a difference in the candidates.
The differences traced back to eleven causes in the simulator, such as the order in which the effects of one card resolve, the content of a few card-reward pools and the counters of a few relics.
After these were fixed, 988 runs matched; after two decisions that the game asks and the simulator had skipped were added (whether to take the relic of a treasure chest, and the map screen that opens each later act), 999 of 1,000 runs matched with observation format~23 and 994 with format~21.
Training and the paired win-rate evaluations below used \href{https://github.com/iRyougi/sts2-sim/tree/v0.111.0.4}{sts2-sim v0.111.0.4} for game v0.111.0 (upstream commit \texttt{abfac840c184}).
The remaining differences were found and fixed while the training was running: the game runs the hooks of a player's powers before those of the player's relics, where the simulator had the opposite order, and three cards and relics differed in a detail.
With these fixes, all 1,000 runs match at every decision with both observation formats; the training run did not use them.
Played greedily in the training build on the same 1,000 seeds, the network of Table~\ref{tab:main-results} won 702 runs, against 710 in the game on the same machine (a paired difference of $-0.8$ points, 95\% interval $-1.9$ to $+0.3$; 11 seeds won only in the simulator and 19 only in the game, exact McNemar $p=0.20$), with the same outcome on 970 seeds.

\paragraph{Training.}
The run follows the Silent's training as closely as its records allow: the configuration with which that training started, and every later change of its schedule (Appendix~\ref{sec:app-recipe}) applied at the update at which it was applied there, including the change of the observation format from 21 to 23 at update 19,901 and the skipping of fights from update 24,208.
It differs from the original training in four ways besides the environment.
First, it ran on one machine with four RTX~5090 GPUs, one for the learner and three for the actors' inference, and 1,344 engines instead of 336; it collected about 15,000 to 16,500 decisions per second, against the 5,000--10,000 of Section~\ref{sec:training}.
Second, it used the released training code for v0.111.0 (Appendix~\ref{sec:code}), not the code with which the original training started.
Third, two episodes of the original training were not replayed: the 517 updates whose weights were discarded, and the 250-update trial of a skipping probability of 1.0 that was then reverted (Appendix~\ref{sec:app-recipe}).
Fourth, the initial outcome heads for the skipping of fights were trained in a 25-update trial on this run's own checkpoint, as in the original training, rather than copied from it.
The random seeds of the training games also differ, as they would between any two runs.

\begin{table}[!ht]
\centering
\caption{Training the Silent in sts2-sim: greedy win rates (\%) of the same checkpoints played in sts2-sim and in the game (v0.111.0) on the same seeds. Raw weights except in the last row. Runs won out of 384 (routine evaluation) or 1,000 (W1) are given in parentheses.}
\label{tab:sim-training}
\small
\begin{tabular}{rrcccc}
\toprule
 & & \multicolumn{2}{c}{Routine evaluation, 384 seeds} & \multicolumn{2}{c}{W1, first 1,000 seeds} \\
Update & Decisions (billions) & sts2-sim & game & sts2-sim & game \\
\midrule
5,000 & 0.16 & 0.5 (2) & 0.5 (2) & 0.6 (6) & 0.5 (5) \\
10,000 & 0.33 & 13.3 (51) & 14.1 (54) & 15.0 (150) & 14.7 (147) \\
15,000 & 0.49 & 29.2 (112) & 29.2 (112) & 28.2 (282) & 28.4 (284) \\
20,000 & 0.70 & 54.7 (210) & 53.1 (204) & 53.3 (533) & 51.7 (517) \\
25,000 & 1.14 & 67.2 (258) & 66.9 (257) & 68.7 (687) & 66.7 (667) \\
26,420 (EMA) & 1.32 & -- & -- & 71.6 (716) & 69.3 (693) \\
\bottomrule
\end{tabular}
\end{table}

\paragraph{Results.}
The routine evaluation used the raw weights and greedy play on 384 seeds every 500 updates. Paired readings in both engines are available from update 1,000 onward. Every 5,000 updates, the raw weights were also evaluated on the first 1,000 seeds of W1 (Table~\ref{tab:sim-training}).
The two evaluations of a checkpoint differ only in the engine that plays the run.
The win rates in the two engines were close, with sts2-sim slightly higher.
Over the 51 readings of the routine evaluation up to update 26,000, the win rate in sts2-sim was on average 0.8 points higher than in the game and differed from it by at most 4.4 points (229 against 212 of 384 runs, at update 22,500); the two engines gave the same outcome on 96.2\% of the runs, and of the others 451 were won only in sts2-sim and 300 only in the game.
The network trained in the game, by contrast, won slightly more often in the game than in sts2-sim (710 against 702, above).
Each network thus did a little better in the engine in which it was trained; we have not investigated why.
As in the original training, the largest gain within a few hundred updates followed the first halving of the learning rate and doubling of the batch (update 18,638): the win rate in the game rose from 35.4\% at update 18,500 to 47.1\% at update 19,000.
At update 26,420, the EMA weights won 69.3\% (95\% Wilson interval 66.4--72.1) of the first 1,000 seeds of W1 in the game and 71.6\% (68.7--74.3) in sts2-sim. All runs finished in both engines. Outcomes agreed on 935 seeds. Of the remaining 65, 44 were won only in sts2-sim and 21 only in the game (exact McNemar $p=0.006$). The network of Table~\ref{tab:main-results}, trained in the game for about the same number of decisions, won 71.3\%. Training reached the simulator checkpoint 25.3 hours after the first update, pauses included, against 2.7 days for the original run.

\paragraph{What this does and does not show.}
The run is one training run with one character on one version of the game, with the differences listed above, so it does not measure how much a run in the simulator and a run in the game differ beyond the variation between two runs in the same engine, which we did not measure either (Section~\ref{sec:discussion}, Limitations).
The lock-step test covers the decisions that one network makes on 1,000 seeds; cards, relics and events that this network rarely or never meets are tested little or not at all.
The fixes were made on the v0.111.0 line of sts2-sim and also merged into its current version, for v0.113.1, which was not checked against the game in this way.

\end{document}